\documentclass[10pt,journal,compsoc]{IEEEtran}

\ifCLASSOPTIONcompsoc
  \usepackage[nocompress]{cite}
\else
  \usepackage{cite}
\fi

\usepackage{epsfig}
\usepackage{graphicx}
\usepackage{amsmath}
\usepackage{amssymb}
\usepackage{graphicx}
\usepackage{amsmath}
\usepackage{amssymb}
\usepackage{algorithm}
\usepackage{algorithmic}
\usepackage{booktabs}
\usepackage{stfloats}
\usepackage{nccmath}
\usepackage{bm}
\usepackage{multirow}
\usepackage{multicol}
\usepackage{array}
\usepackage{url}         
\usepackage{xcolor} 
\usepackage{epsfig} 
\usepackage{colortbl}
\usepackage{arydshln}
\usepackage{array}
\usepackage{booktabs}
\usepackage{threeparttable}
\usepackage[shortlabels]{enumitem}
\usepackage{subcaption}
\usepackage{caption}
\usepackage{soul, color, xcolor}
\usepackage{ulem}

\begin{document}

\newcommand{\tabincell}[2]{\begin{tabular}{@{}#1@{}}#2\end{tabular}}
\newcommand{\PreserveBackslash}[1]{\let\temp=\\#1\let\\=\temp}
\newcommand{\eg}{\textit{e}.\textit{g}. }
\newcommand{\ie}{\textit{i}.\textit{e}. }
\renewcommand{\figurename}{Figure.}
\newcolumntype{C}[1]{>{\PreserveBackslash\centering}p{#1}}
\definecolor{blue}{rgb}{0.062,0.309,0.725}
%

\title{ImFace++: A Sophisticated Nonlinear 3D Morphable Face Model with Implicit Neural Representations}
 
%
%
%
%

\author{Mingwu~Zheng*,~\IEEEmembership{Student Member,~IEEE,}
        Haiyu~Zhang*,~\IEEEmembership{Student Member,~IEEE,}
        Hongyu~Yang,~\IEEEmembership{Member,~IEEE,} 
        Liming Chen,~\IEEEmembership{Senior Member,~IEEE}
        and Di~Huang$^{\dag}$,~\IEEEmembership{Senior Member,~IEEE,}
\IEEEcompsocitemizethanks{\IEEEcompsocthanksitem M. Zheng, H. Zhang and D. Huang are with the State Key Laboratory of Complex and Critical Software Environment, School of Computer Science and Engineering, Beihang University, Beijing 100191, China. E-mail:\{zhengmingwu, zhyzhy, dhuang\}@buaa.edu.cn
\IEEEcompsocthanksitem H. Yang is with the School of Artificial Intelligence, Beihang University, Beijing 100191, China. She is also with Shanghai Artificial Intelligence Laboratory, Shanghai 201112, China.  E-mail: hongyuyang@buaa.edu.cn.

\IEEEcompsocthanksitem L. Chen is with the Department of Mathematics and Informatics, Ecole Centrale de Lyon, University of Lyon, France. E-mail: liming.chen@ec-lyon.fr.

\IEEEcompsocthanksitem *Equal contribution. $^{\dag}$Corresponding author.}
}

\IEEEtitleabstractindextext{%
\begin{abstract}

Accurate representations of 3D faces are of paramount importance in various computer vision and graphics applications. However, the challenges persist due to the limitations imposed by data discretization and model linearity, which hinder the precise capture of identity and expression clues in current studies. This paper presents a novel 3D morphable face model, named ImFace++, to learn a sophisticated and continuous space with implicit neural representations. ImFace++ first constructs two explicitly disentangled deformation fields to model complex shapes associated with identities and expressions, respectively, which simultaneously facilitate automatic learning of point-to-point correspondences across diverse facial shapes. To capture more sophisticated facial details, a refinement displacement field within the template space is further incorporated, enabling fine-grained learning of individual-specific facial details. Furthermore, a Neural Blend-Field is designed to reinforce the representation capabilities through adaptive blending of an array of local fields. In addition to ImFace++, we devise an improved learning strategy to extend expression embeddings, allowing for a broader range of expression variations. Comprehensive qualitative and quantitative evaluation demonstrates that ImFace++ significantly advances the state-of-the-art in terms of both face reconstruction fidelity and correspondence accuracy. 

\end{abstract}

}

\maketitle

\IEEEdisplaynontitleabstractindextext

%
\IEEEpeerreviewmaketitle

\IEEEraisesectionheading{\section{Introduction}\label{sec:introduction}}
\IEEEPARstart{3}{D} Morphable Face Models (3DMMs) are well-reputed statistical models, established by learning techniques upon prior distributions of facial shapes and textures from a set of samples with dense correspondence, aiming at rendering realistic faces of a high variety. Since a morphable representation is unique across different downstream tasks where the geometry and appearance are separately controllable, 3DMMs are pervasively exploited in many face analysis applications in the field of computer vision, computer graphics, biometrics, and medical imaging~\cite{aldrian2012inverse, blanz2003face, hu2016face, staal2015describing}.

In 3DMMs, the most fundamental issue lies in the way to generate latent morphable representations, and during the past two decades, along with data improvement in scale, diversity, and quality~\cite{cao2013facewarehouse, booth2018large, li2017flame, yang2020facescape}, remarkable progresses have been achieved. The methods are initially linear model based~\cite{blanz1999morphable, patel20093d, bfm09} and further multilinear model based~\cite{vlasic2006face, brunton2014multilinear, bolkart2015groupwise}, where different modes are individually encoded. Unfortunately, for the relatively limited representation ability of linear models, these methods are not so competent at handling the cases with complex variations, \eg, exaggerated expressions. In the era of deep learning, a number of nonlinear models have been investigated with the input of 2D images~\cite{tran2018nonlinear, tran2019towards} or 3D meshes~\cite{bagautdinov2018modeling, ranjan2018generating, bouritsas2019neural, chen2021learning, cheng2019arxiv} by using Convolutional Neural Networks (CNN) or Graph Neural Networks (GNN). They indeed deliver performance gains; however, restricted by the resolution of discrete representing strategies on input data, facial priors are not sufficiently captured, incurring loss of shape details. Besides, all current methods are dependent on the preposed procedure of point-to-point correspondence~\cite{gilani2017dense, abrevaya2018spatiotemporal, liu20193d, bahri2021shape}, but face registration itself remains challenging. 

\begin{figure*}
  \centering
   \setlength{\abovecaptionskip}{0pt}
   \setlength{\belowcaptionskip}{0pt}
   \includegraphics[width=1\linewidth]{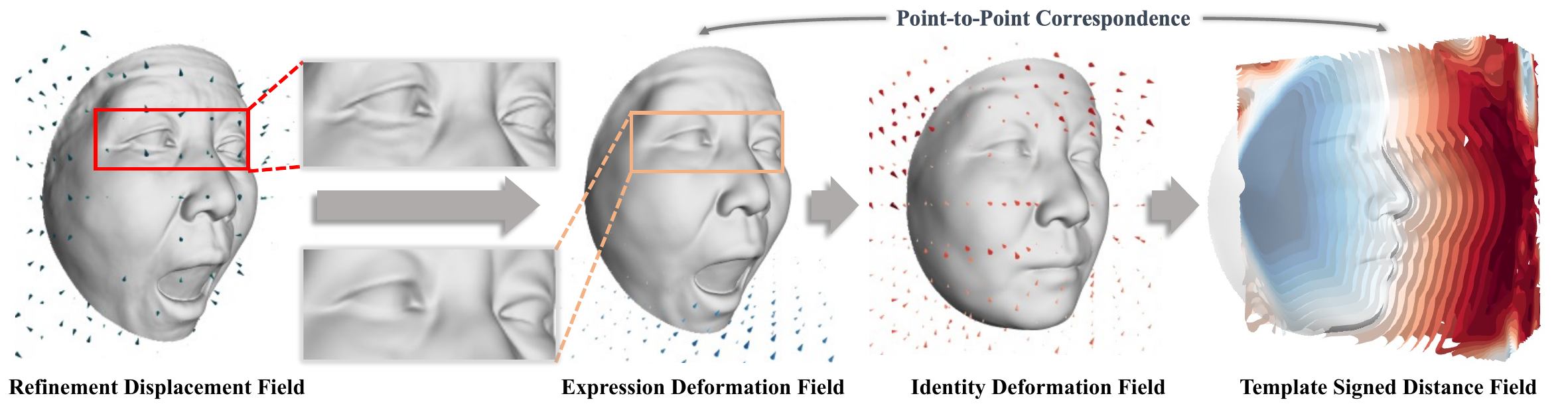}
   \caption{ImFace++ encodes intricate variations associated with identities and expressions by two explicitly disentangled deformation fields with respect to a template face. Leveraging the established point-to-point correspondence, it applies an additional refinement displacement field to capture high-frequency details in the individual-independent template space. The model finally yields sophisticated and implicit representations that are highly adaptable for 3D face modeling.} 
   \label{fig:sample}
\end{figure*}

Recently, several studies have illuminated the potential of Implicit Neural Representations (INRs) \cite{park2019deepsdf, mescheder2019occupancy, chen2019learning, icml2020_2086, pmlr-v139-lipman21a} in precise modeling of 3D geometries. INRs describe input observations as low-dimensional shape embeddings and estimate the Signed Distance Function (SDF) or the occupancy values of query points, allowing for the definition of surfaces with arbitrary resolutions and topologies through isocontours. Such continuous representations offer advantages over the traditional discrete ones like voxels, point clouds, and meshes, and have reported substantial successes in shape reconstruction~\cite{xu2019disn, wang2021neus, genova2019iccv,zhang2021learning} as well as surface registration~\cite{deng2021deformed, liu2020nips, zheng2021deep}. This property suggests an alternative to 3DMM that can fulfill accurate shape correspondence and geometry modeling in a unified network. Despite that, applying INRs to face modeling presents unique challenges. Unlike the objects with distinct shapes and limited non-rigid variations, such as indoor scenes and human bodies, facial surfaces are inherently similar but also encompass intricate deformations arising from the interweaving of multiple identities and rich expressions. Current INR methods struggle to precisely capture and disentangle the intricacies, as evidenced by the preliminary attempt ~\cite{yenamandra2021i3dmm}.

On the other side, it is essential to emphasize the faithful reproduction of facial details.  In the context of geometry modeling, there has been a growing interest in hybrid approaches that combine explicit and implicit representations to harness the strengths of both techniques~\cite{muller2022instant,chan2022efficient,sun2022direct}, aiming to address the spectral bias~\cite{rahaman2019spectral} issue often observed in neural networks and learn more sophisticated geometry clues.
However, incorporating explicit 3D representations into INRs can introduce discontinuities, potentially leading to training instability and over-fitting~\cite{mihajlovic2023resfields,zhao2024grounding}, especially when applied to complex structures like facial surfaces. 
An alternative research trend focuses on incorporating computer graphics techniques,  such as displacement mapping~\cite{yifan2021geometry, wang2022hf}, into INRs to encode rich details. While these methods have shown remarkable performance in reconstructing single static scenes, the challenge remains in developing INR models capable of effectively capturing dynamically changing high-frequency details across multiple facial shapes, such as wrinkles and dimples.

In response to the aforementioned challenges, this paper presents a novel two-stage 3D face morphable model, namely ImFace++, which substantially upgrades conventional 3DMMs by integrating customized INRs. In its initial stage, ImFace++ tackles nonlinear facial deformation modeling and establishment of point-to-point correspondence among different facial shapes by creating separate INR sub-networks, which explicitly disentangle shape morphs into two deformation fields for identity and expression, respectively (as Fig.~\ref{fig:sample} shows). The template shape is defined by a deep SDF, therefore eliminating the need for the face registration phase required by numerical solutions. 
To build more sophisticated 3DMMs, ImFace++ further incorporates a Refinement Displacement Field (RDF) in the second stage to capture high-frequency geometry details present in various faces. Notably, this stage for detail modeling is specifically tailored to adapt to the dynamic nature of facial expressions. By leveraging the established correspondences, RDF is learned within the individual-independent template space characterized by its relative stability, allowing the model to concentrate on encoding subtle details. Overall, by learning the disentangled deformation fields and RDF, ImFace++ accurately represents inter-individual changes and fine-grained clues while retaining the flexibility for its application in related tasks.

Inspired by linear blend skinning~\cite{lewis2000pose}, we further introduce a Neural Blend-Field to decompose the entire facial deformation or geometry into semantically meaningful regions encoded by an array of local implicit functions and adaptively blend them through a lightweight module, leading to more advanced representations with reduced parameter overheads. In addition to ImFace++, an improved embedding learning strategy is devised to extend the latent space of expressions to allow a more diverse range.

Overall, this study makes the following contributions:
\begin{itemize}
\item A novel INR-based 3DMM, namely ImFace++. It encodes complex facial shape variations with identity and expression through explicitly disentangled deformation fields, facilitating learning flexible and semantically meaningful representations. 

\item A refinement displacement field, \ie RDF, within the individual-independent template space. Leveraging the established correspondence, the field faithfully encodes high-frequency facial details, thus enhancing the representation capabilities of ImFace++.

\item A Neural Blend-Field. It enables fine-grained learning by adaptively blending an array of local implicit functions.

\item Compelling results in different tasks. Extensive experiments evidence that ImFace++ excels in synthesizing high-quality 3D faces with intricate details, outperforming the state-of-the-art alternatives.

\end{itemize}
 
A preliminary version of this study, \ie ImFace, previously published in~\cite{zheng2022imface}. This paper significantly improves it in the following aspects: 
(i) We extend the model to a two-stage framework. To sufficiently capture facial priors, the upgraded model leverages the correspondences between different individuals, achieved by employing disentangled deformation fields in the first stage as in ImFace~\cite{zheng2022imface}, and further applies a newly designed refinement field to progressively model face geometry, significantly alleviating the loss of shape details while relieving the burden of model learning.
(ii) We update the model to reconstruct and synthesize 3D faces spanning a wide range of facial expressions and it renders more vivid results. We also refine the evaluation in both the observed and latent representation spaces, along with more comprehensive discussions. 
(iii) To further validate the proposed approach, we conduct comparison with other INR-based methods~\cite{yenamandra2021i3dmm,giebenhain2022learning} in terms of 3D face reconstruction and face correspondence.
(iv) We showcase applications in various scenarios, including facial expression editing, detail transfer, multi-view reconstruction, and ear-to-ear face reconstruction, to demonstrate the adaptability of ImFace++ in the realm of 3D face modeling.

The rest of this paper is organized as follows. Sec. 2 reviews related work on 3DMMs and INRs. Sec. 3 elaborates on the proposed ImFace++ method. Sec. 4 presents and analyzes the experimental results, followed by Sec. 5, which concludes this paper and provides insights for future perspectives.

\section{Related Work}

The published studies closely related to our study can be primarily summarized into: (i) 3D Morphable Models (3DMMs), (ii) Implicit Neural Representations (INRs), in particular those aiming to capture complex 3D structures, and (iii) fine-grained neural representations for geometry modeling. In the following, we provide a concise review of the relevant literature.

\subsection{3D Morphable Face Models} 

The concept of 3DMMs was first introduced by Blanz and Vetter \cite{blanz1999morphable} as a foundation framework for face representation. This approach involves registering a known template mesh to all training scans using the Non-rigid Iterative Closest Point (NICP) algorithm \cite{amberg2007optimal}. Principal Components Analysis (PCA) is subsequently employed to create prior face distributions. To account for identity-dependent expressions, 3DMMs were extended to multilinear models \cite{vlasic2006face, brunton2014multilinear, bolkart2015groupwise}. Over time, great progress has been made aided by improvements in data quality \cite{cao2013facewarehouse, booth2018large, li2017flame, yang2020facescape}. Notably, the Facial LAndmarks and More Expressions (FLAME) model \cite{li2017flame} was introduced as an expressive model, capable of controlling facial expressions through a combination of jaw articulation and linear expression blendshapes. Trained on a comprehensive 3D facial dataset, including D3DFACS \cite{cosker2011facs}, FLAME delivers remarkable results, although it faces challenges in accurately capturing nonlinear facial deformations.

Deep neural networks have significantly accelerated the development of more powerful nonlinear 3DMMs. A number of models have been trained directly from 2D images \cite{tran2018nonlinear, tran2019towards, R_2021_CVPR, shi2020neural}, but they often struggle to capture high-fidelity and fine-grained details. Such a limitation is attributed to the inherent challenges posed by the low resolution of input images in solving this ill-posed inverse problem. Recognizing the need to better leverage 3D face scans, \cite{bagautdinov2018modeling} devises a method for mapping 3D meshes to the 2D space. Subsequently, several studies \cite{cheng2019arxiv, ranjan2018generating, chen2021learning, bouritsas2019neural} pursue the learning of 3DMMs from meshes, employing spectral or spiral convolutions. ASM~\cite{yang2023asm} presents an adaptive skinning model, which utilizes the Gaussian Mixture Model (GMM) to generate skinning weights. While promising, these neural networks are established on discrete 3D representations, which tend to fail when confronting with intricate facial deformations and high-frequency details.

For a more comprehensive survey on 3DMMs, readers are encouraged to refer to \cite{egger20203d}.

\subsection{Implicit Neural Representations} 

In recent years, implicit neural functions have emerged as powerful and effective representations for 3D geometries \cite{park2019deepsdf, mescheder2019occupancy, chen2019learning, icml2020_2086, pmlr-v139-lipman21a, sitzmann2019siren}. These approaches revolutionize the field by enabling continuous and highly expressive representations of geometric structures without the need for discretization. To address the complexity of shape variations and correspondence relationships, some attempts further incorporate an additional implicit deformation latent space \cite{deng2021deformed, zheng2021deep}.

A number of studies focus on applying INRs to modeling human heads or bodies~\cite{corona2021smplicit,ramon2021h3d,yenamandra2021i3dmm,Alldieck_2021_ICCV,Chen_2021_ICCV,peng2021neural,saito2019pifu,saito2020pifuhd, giebenhain2022learning,wang2022morf,hong2022headnerf,zhuang2022mofanerf, caselles2023implicit, buhler2023preface, palafox2021npms, mohamed2022gnpm}. In particular, NPMs~\cite{palafox2021npms} and GNPM~\cite{mohamed2022gnpm} employ two neural networks to separately model the body shape and pose spaces. Models such as H3D-Net~\cite{ramon2021h3d}, SIDER~\cite{chatziagapi2021sider}, and SIRA++\cite{caselles2023implicit} learn implicit priors for 3D head reconstruction. i3DMM~\cite{yenamandra2021i3dmm} makes significant strides by pioneering a generative implicit 3DMM tailored for human heads. However, utilizing neural networks to generate fine-grained details from raw signals often results in over-smoothed or implausible facial shapes. A concurrent work, NPHM~\cite{giebenhain2022learning}, introduces a parametric head model using local implicit fields, and it adopts local INRs for identity representation and a globally conditioned INR for modeling expression deformations. Nonetheless, its ability for capturing high-frequency facial details is constrained by the globally conditioned neural network architecture. Another series of works~\cite{wang2022morf,hong2022headnerf,zhuang2022mofanerf, buhler2023preface} concentrate on modeling facial appearances, leveraging techniques like differential volume rendering \cite{mildenhall2020nerf} and auto-decoding \cite{park2019deepsdf} to learn morphable radiance fields. But these models represent the underlying geometry using density, leading to noisy geometry approximations.

\begin{figure*}
  \centering
  \includegraphics[width=1\linewidth]{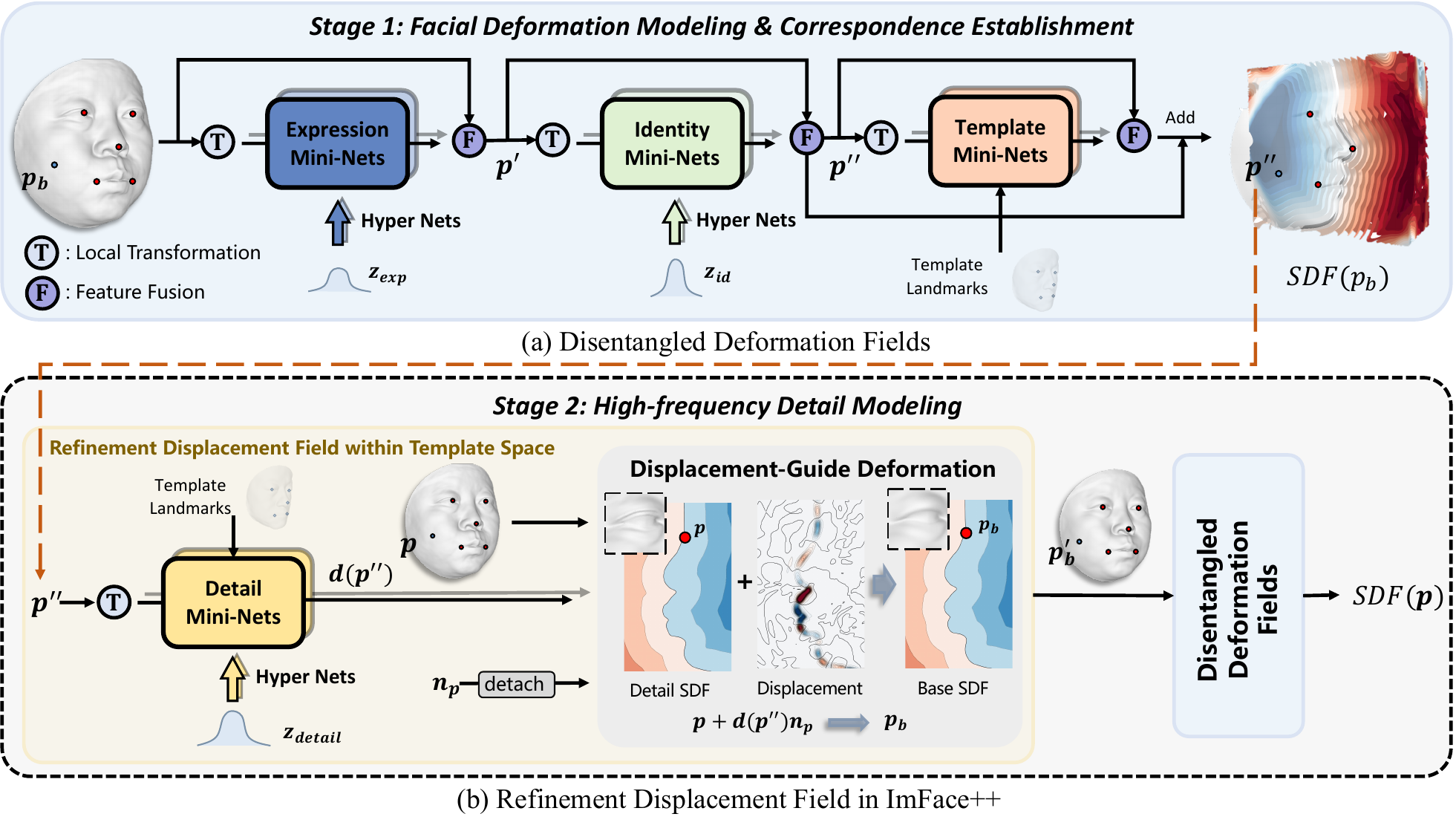}
  \caption{{\bf ImFace++ overview.} The model is constructed through two-stage learning. \textbf{(a)} The first stage primarily focuses on nonlinear deformation modeling and surface correspondence between different faces. The network comprises three Mini-Nets blocks designed to explicitly disentangle shape morphs into separate deformation fields, where the Expression and Identity Mini-Nets blocks are associated with expression and identity deformations, respectively. Meanwhile, the Template Mini-Nets block learns the SDF of a template face shape. \textbf{(b)} With the point-to-point correspondence established, ImFace++ proceeds to employ a Detail Mini-Nets block in the second stage, responsible for learning refinement displacement field that encodes high-frequency geometry details.
  }
  \label{fig:pipeline}
\end{figure*}

\subsection{Fine-Grained Neural Representations for Geometry Modeling}

Recent research~\cite{rahaman2019spectral,xu2019training,basri2020frequency} brings attention to a persistent challenge in neural networks that the learned models tend to favor the retention of low-frequency signals while neglecting high-frequency information, known as the spectral bias~\cite{rahaman2019spectral}. In the context of geometry modeling, latest advancements introduce hybrid approaches that combine implicit and explicit 3D representations, \eg voxels~\cite{sun2022direct,martel2021acorn,wu2022voxurf}, tri-planes~\cite{chan2022efficient,shue20233d} and hash-grids~\cite{muller2022instant,xu2022manvatar,gao2022reconstructing}, to mitigate the issue above. In particular, Voxurf~\cite{wu2022voxurf} employs voxel-based representations combined with an implicit function defined by a shallow MLP to represent a scene. It also builds hierarchical geometry features to enhance perception across voxels. EG3D~\cite{chan2022efficient} presents a memory-efficient tri-plane architecture that efficiently scales with resolution, yielding a higher capacity for handling complex 3D data. TensoRF~\cite{chen2022tensorf} decomposes the 3D volume into vector-matrix outer products, leading to improved rendering quality. Additionally, INGP~\cite{muller2022instant} combines voxel grids and multi-resolution hash tables to create compact and adaptive data representations. This approach results in substantial enhancements in both rendering quality and processing speed. 
Inspired by the domain of signal processing, NeuRBF~\cite{chen2023neurbf} and NFFB~\cite{wu2023neural} utilize Radial Basis Functions (RBF) and filter banks as irregular grids to improve the capability. However, using these hybrid 3D representations as 3DMMs is challenging due to discontinuities in explicit 3D representations, incurring inaccurate gradients and poor generalizability, as discussed in the supplementary material of iNGP~\cite{muller2022instant} and \cite{zhao2024grounding}.

Another research trend seeks to bolster the intrinsic capabilities of neural networks for improved details. Inspired by the natural language processing field, NeRF~\cite{mildenhall2020nerf} applies position encoding to render finer-grained details. In addition, SIREN~\cite{sitzmann2020implicit}, GARF~\cite{chng2022gaussian}, and WIRE~\cite{saragadam2023wire} introduce various network activation functions, including Sin, Gaussian, and Gabor wavelet functions. However, training these networks is not so straightforward due to instability and susceptibility to over-fitting. To overcome this, FreeNeRF~\cite{yang2023freenerf} and Nerfies~\cite{park2021nerfies} add a coarse-to-fine learning strategy. Furthermore, IDF~\cite{yifan2021geometry} and HF-NeuS~\cite{wang2022hf} extend the conventional explicit and discrete displacement mapping into the ${\mathbb R}^3$ domain to capture high-frequency details within single static scenes. Nevertheless, there is relatively limited exploration into the modeling of dynamic high-frequency facial details across different individuals. To enhance the ability of INRs for 3DMMs, our method employs a decomposition of the learning process, which begins by automatically learning deformations linked to identity and expression while simultaneously establishing correspondences between different individuals. It then learns to encode dynamic high-frequency details in the established template space, significantly facilitating the representation capability.

\section{Method}

We take advantages of INRs to learn a sophisticated nonlinear 3DMM. The fundamental idea of INRs is to train a neural network to fit a continuous function $f$, which implicitly represents surfaces through level-sets. The function can be defined in various formats, \eg,  occupancy~\cite{mescheder2019occupancy}, SDF~\cite{park2019deepsdf}, or UDF~\cite{chibane2020ndf}. We exploit a deep SDF conditioned on the latent embeddings associated with various facial characteristics for comprehensive facial representations. It outputs the signed distance $s$ from a query point:
\small{
\begin{equation}
  f:({\mathbf p}, {\mathbf z}_{exp}, {\mathbf z}_{id}, {\mathbf z}_{detail}) \in {\mathbb R}^{3+d_{exp}+d_{id}+d_{detail}} \mapsto  s \in {\mathbb R},
\end{equation}
}

\noindent where ${\mathbf p} \in {\mathbb R}^3 $ is the coordinate of the query point in the 3D space, ${\mathbf z}_{exp}$, ${\mathbf z}_{id}$, and ${\mathbf z}_{detail}$ denote the expression, identity and high-frequency detail embeddings of a given face, respectively.

Our goal is to learn a neural network to parameterize $f$, making it satisfy the genuine facial shape priors. As shown in Fig.~\ref{fig:pipeline}, the proposed network for ImFace++ is constructed by a series of Mini-Nets blocks that are progressively learned. In the first stage, ImFace++ employs the Expression and Identity Mini-Nets blocks to learn separate deformation fields associated with expression- and identity-variations, respectively. Meanwhile, a Template Mini-Nets block is used to learn a signed distance field of a template face shape. Such explicitly disentangled architecture ensures that the intricate expression deformations and inter-individual differences can be accurately modeled (Sec.~\ref{subsec:disentangled}). The second stage complements the former by incorporating a Detail Mini-Nets block to learn a refinement displacement field, \ie RDF, within the template space, which faithfully encodes facial geometry details (see Sec.~\ref{subsec:idf}). The network parameters and facial embeddings are jointly learned. After convergence, a face scan can be disentangled by deforming it from its detailed state to the template. 

All the aforementioned fields are implemented by a shared Mini-Nets architecture, where the entire facial deformation or geometry is decomposed into an array of semantically meaningful parts and encoded by a set of local field functions so that intricate details can be sufficiently captured. A lightweight module conditioned on the query point position, \textit{i.e,} Fusion Network, is stacked at the end of the Mini-Nets block to adaptively blend the local fields. As such, an elaborate Neural Blend-Field is achieved (see Sec.~\ref{subsec:nbf}). Additionally, an improved embedding learning strategy is designed, allowing more diverse details of expressions (refer to Sec.~\ref{subsec:emb}). Detailed explanations of the model designs are provided as follows.

\subsection{Disentangled Deformation Fields}
\label{subsec:disentangled} 

The proposed method initially focuses on non-linear facial deformation modeling and establishment of point-to-point correspondences among different faces. This is accomplished through three Mini-Nets blocks as previously outlined. As Fig.~\ref{fig:pipeline} (a) shows, we collectively refer to this process as \textbf{Disentangled Deformation Fields} modeling and the learned SDF can finally reproduce an expression space (Fig.~\ref{fig:model_analysis} illustrates the differences between these spaces), which encompasses facial shapes with different expressions and identities. The core components serve distinct purposes, with slight structural adaptations as follows: 

\subsubsection{Expression Mini-Nets (ExpNet)} The facial deformations incurred by expressions are represented by ExpNet $\mathcal{E}$, which learns a warping from the expression space to the identity space for each face scan:
\begin{equation}
  \mathcal{E}:({\mathbf p}_{b}, {\mathbf z}_{exp}, l)  \mapsto  {\mathbf p}' \in {\mathbb R}^{3},
  \label{exp_deform}
\end{equation}

\noindent where ${\mathbf p}_{b}$ denotes a query point in the expression space, and $l \in {\mathbb R}^{k \times 3}$ denotes $k$ 3D landmarks on faces in the expression space generated by a Landmark-Net $\eta: ({\mathbf z}_{exp}, {\mathbf z}_{id}) \mapsto l $, which is introduced to localize the query point ${\mathbf p}_{b}$ in the Neural Blend-Field. This point is deformed by $\mathcal{E}$ to ${\mathbf p}'$ in the person-specific identity space, which represents faces with a neutral expression.

\subsubsection{Identity Mini-Nets (IDNet)} To model shape morphs among individuals, IDNet $\mathcal{I}$ further warps the identity space to a template shape space shared by all faces:
\begin{equation}
  \mathcal{I}:({\mathbf p}', {\mathbf z}_{id}, l')  \mapsto  ({\mathbf p}'', \delta) \in {\mathbb R}^{3} \times {\mathbb R},
  \label{id_deform}
\end{equation}

\noindent where $l' \in {\mathbb R}^{k \times 3}$ denotes $k$ landmarks on a face with a neutral expression generated by another Landmark-Net conditioned only on the identity embedding, namely  $\eta': {\mathbf z}_{id} \mapsto l' $, where ${\mathbf p}''$ is the deformed point in the template space. To cope with the possible non-existent correspondences produced during preprocessing, $\mathcal{I}$ additionally predicts a residual term $\delta \in {\mathbb R}$ to correct the SDF value $s_0$ predicted by the subsequent TempNet, which is similar to \cite{deng2021deformed}.

\subsubsection{Template Mini-Nets (TempNet)} TempNet $\mathcal{T}$ learns a signed distance field of the shared template face:
\begin{equation}
  \mathcal{T}:({\mathbf p}'', l'')  \mapsto  s_0 \in {\mathbb R},
\end{equation}
where \noindent $l'' \in {\mathbb R}^{k \times 3}$ denotes $k$ landmarks on the template face, which is averaged on the whole training set, and $s_0$ is the uncorrected SDF value. The final SDF value of a query point is calculated via $s = s_0 + \delta$, and the modeling process of disentangled deformation fields can be collectively formulated as:
{\small
\begin{equation}
  \hat{f}({\mathbf p}_{b}, {\mathbf z}_{exp}, {\mathbf z}_{id}) = \mathcal{T}(\mathcal{I}(\mathcal{E}({\mathbf p}_{b}))) + \mathcal{I}_{\delta}(\mathcal{E}({\mathbf p_{b}})).
\end{equation}
}

\noindent For brevity, the latent embeddings and predicted landmarks are omitted. $\mathcal{I}_{\delta}(\mathcal{E}({\mathbf p_{b}}))$ denotes the residual term $\delta$ in Eq.~(\ref{id_deform}). By doing so, we can automatically learn correspondence across different individuals by mapping the query point ${\mathbf p}_{b}$ to the template point ${\mathbf p''}$.

\begin{figure*}
  \centering
  \setlength{\abovecaptionskip}{0pt}
  \setlength{\belowcaptionskip}{0pt}
  \includegraphics[width=1.0\linewidth]{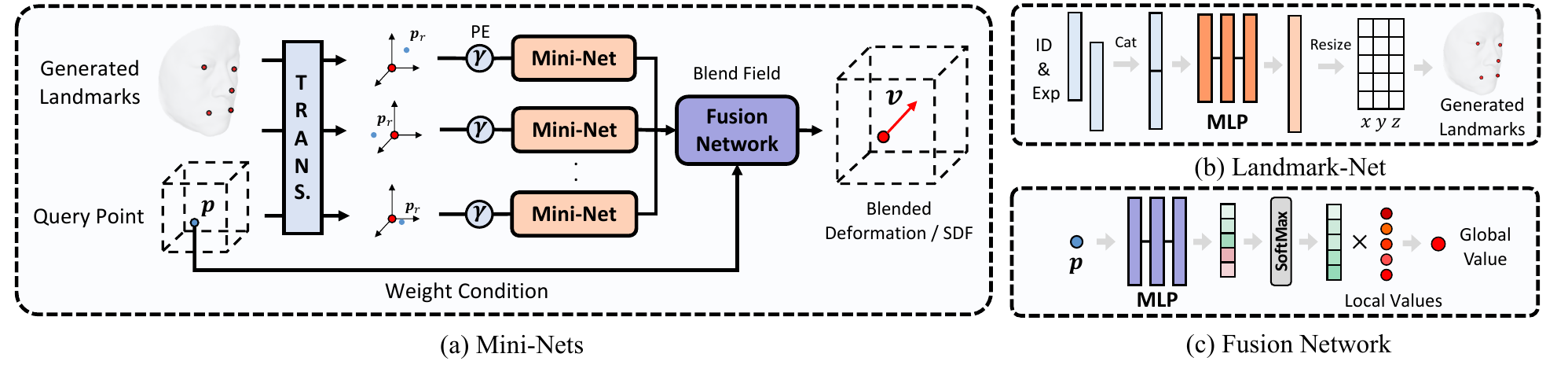}
   \caption{\textbf{Neural Blend-Field.} \textbf{(a)} The Mini-Nets block is a shared architecture, which decomposes an entire facial feature into semantically meaningful parts and encodes them by a set of local field functions. It is tailed by a Fusion Network for more comprehensive representations. \textbf{(b)} The Landmark-Net is introduced to softly partition the entire facial surface. \textbf{(c)} The Fusion Network is a lightweight module conditioned on the query point position, which adaptively blends the local field functions, resulting in an elaborate Neural Blend-Field.}
   \label{fig:mini-nets}
\end{figure*}

\subsection{Refinement Displacement Field}

\label{subsec:idf}

Facial modeling assumes a pivotal role in enhancing the overall visual fidelity and realism of facial representations. The intricate nuances, such as fine lines, wrinkles, and subtle expressions, collectively contribute to the creation of a truly lifelike and convincing depiction of a face. To construct an advanced 3DMM, we introduce a specialized refinement module designed to master high-frequency facial details. Drawing inspiration from the concept of Implicit Displacement Fields (IDF) proposed in~\cite{yifan2021geometry}, where high-frequency details are generated by offsetting a smooth base surface along its normal directions based on the displacement distances predicted from IDF, we develop a Detail Mini-Nets block (DetailNet), denoted as $\mathcal{D}$, to learn an RDF, as depicted in Fig.~\ref{fig:pipeline} (b). To account for the dynamic nature of human faces, instead of representing facial geometry details using a conventional IDF, we encode them in the individual-independent template space that exhibits a relative stability so that subtle changes can be well captured. This strategy is reminiscent of the concept of storing displacement information in a topology-consistent UV texture map.

To achieve this, we leverage the correspondences established during disentangled deformation field modeling and RDF is learned in the later stage. 
Formally, its modeling process can be expressed as:

\begin{equation}
  \mathcal{D}:({\mathbf p}'', {\mathbf z}_{detail}, l'')  \mapsto  {\mathbf d} \in {\mathbb R},
\end{equation}

\noindent where ${\mathbf d}$ denotes the displacement distance along the normal of a point ${\mathbf p}$, which tends to be overlooked in the first learning stage. It is worth noting that ${\mathbf p}''$ is the input of this function, ensuring that the subtle displacements are learned in the template space. Such learning phase decomposition effectively isolates subtle displacements from identity and expression-related deformations, and the facial details can be readily applied to other facial shapes that share the same template space. Taking the detail displacement into consideration, the coordinates of ${\mathbf p}_{b}$ in the aforementioned expression space represented by $\hat{f}$ can be corrected as follows:
{\small
\begin{equation}
   {\mathbf p}'_{b}= {\mathbf p} + \chi(\hat{f}({\mathbf p}))\mathcal{D}({\mathbf p}'', {\mathbf z}_{detail}, l'') \frac{\nabla \hat{f}({\mathbf p})}{||\nabla \hat{f}({\mathbf p})||},
   \label{detail_deform}
\end{equation}
}

\noindent where $\chi$ denotes the attenuation function~\cite{yifan2021geometry} and $\mathbf{p}''=\mathcal{I}(\mathcal{E}(\mathbf{p}_{b}))$. $\frac{\nabla \hat{f}({\mathbf p})}{||\nabla \hat{f}({\mathbf p})||}$ denotes the normalized normal of ${\mathbf p}$ calculated in the expression space. This approach accounts for the influence of expressions when modeling facial details and ensures that geometries not captured by the disentangled deformation fields are fully learned. Finally, for a face scan, its key components (\textit{i.e.,} expression and identity) can be accurately disentangled by performing the following warping process: 
{\small
\begin{equation}
  f({\mathbf p}, {\mathbf z}_{exp}, {\mathbf z}_{id}, {\mathbf z}_{detail}) = \mathcal{T}(\mathcal{I}(\mathcal{E}({\mathbf p}'_{b}))) + \mathcal{I}_{\delta}(\mathcal{E}({\mathbf p}'_{b})).
  \label{warping_process}
\end{equation}
}

Notably, while Eq.~(\ref{warping_process}) ideally requires \(\mathbf{p}_{b}\) as input, obtaining exact matching pairs (\(\mathbf{p}_{b}\), \(\mathbf{p}\)) in the 3D space is difficult as the facial shapes are typically provided in explicit form like meshes.
For practical implementation, we approximate \(\mathbf{p}'' = \mathcal{I}(\mathcal{E}(\mathbf{p}))\), considering the small displacement distance. The proposed ImFace++ learns face morphs by disentangled deformation fields and RDF in a fine-grained and meaningful manner, ensuring that more diverse and sophisticated facial deformations, correspondence, and details can be accurately captured.

\subsection{Neural Blend-Field}
\label{subsec:nbf}

The Mini-Nets block is a common architecture shared by the sub-networks $\mathcal{E}$, $\mathcal{I}$, $\mathcal{T}$, and $\mathcal{D}$. As shown in Fig.~\ref{fig:mini-nets}, it learns a continuous field function ${\boldsymbol \psi}:{\mathbf x} \in {\mathbb R}^3 \mapsto v $ (\eg, deformation, displacement distance or signed distance value), to produce a Neural Blend-Field for comprehensive face representations. In particular, to overcome the limited expressivity of a single network, we decompose a face into a set of semantically meaningful local regions and learn $v$ individually before blending. Such design is inspired by the recent INRs study~\cite{peng2021animatable} on human body, which introduces the linear blend skinning algorithm~\cite{lewis2000pose} to make the network learn from separate transformations of body parts. To better represent detailed facial shapes, we replace the constant transformation term in the original linear blend skinning algorithm with ${\boldsymbol \psi}_n({\mathbf x}-l_n)$, and define the Neural Blend-Field as:

\begin{equation}
  v={\boldsymbol \psi}({\mathbf x}) =\sum_{n=1}^k{{\it w}_n({\mathbf x}) {\boldsymbol \psi}_n({\mathbf x}-l_n)},
\end{equation}
\noindent where $l_n$ is a parameter that describes the $n$-th local region, ${\it w}_n({\mathbf x})$ is the $n$-th blend weight, ${\boldsymbol \psi}_n({\mathbf x}-l_n)$ is the corresponding local field, and $k$ is the number of local regions. In this way, the blending is performed on a series of local fields, rather than calculating a weighted average of the output values $v$ of some fixed positions, leading to a more powerful representation capability in handling complicated local features.

Specifically, five landmarks located at the outer eye corners, mouth corners, and nose tip are utilized to describe the local regions $(l_n \in {\mathbb R}^{3})_{n=1}^5$, which is predicted by Landmark-Net, as illustrated in Fig.~\ref{fig:mini-nets} (b). As Fig.~\ref{fig:mini-nets} (a) shows, each region is assigned a tiny MLP with sinusoidal activations~\cite{sitzmann2019siren} to generate the local field, denoted as ${\boldsymbol \psi}_n$. To capture high-frequency local variations, we leverage sinusoidal positional encoding $\gamma$~\cite{mildenhall2020nerf} on the coordinate ${\mathbf x}-l_n$. At the end of a Mini-Nets block, a lightweight Fusion Network conditioned on the absolute coordinate of input ${\mathbf x}$ is equipped, which is implemented by a 3-layer MLP with softmax to predict the blend weights $(w_n \in {\mathbb R}^+)_{n=1}^5$, as illustrated in Fig.~\ref{fig:mini-nets} (c). 

\subsubsection{Deformation Formulation} We formulate the deformation with an SE(3) field $({\boldsymbol \omega},{\mathbf v}) \in {\mathbb R}^{6} $, where ${\boldsymbol \omega} \in so(3)$ is a rotate vector representing the screw axis and the angle of rotation. The deformed coordinates ${\mathbf x}'$ can be calculated by ${e^{\boldsymbol \omega}}{\mathbf x}+{\mathbf t}$, where the rotation matrix ${e^{\boldsymbol \omega}}$ (exponential map form of Rodrigues’ formula) is written as:
{\small
\begin{equation}
  {e^{\boldsymbol \omega}}={\bf I}+ \frac{\sin \left \| {\boldsymbol \omega} \right \|}{\left \| {\boldsymbol \omega} \right \|} {\boldsymbol \omega}^{\wedge} + \frac{1-\cos \left \| {\boldsymbol \omega} \right \|}{{\left \| {\boldsymbol \omega} \right \|}^2} ({\boldsymbol \omega}^{\wedge})^2,
\end{equation}}
\hspace{-0.5em} and the translation ${\mathbf t}$ is formulated as:
{\small
  \begin{equation}
    {\mathbf t}=\left[{\bf I}+ \frac{1-\cos \left \| {\boldsymbol \omega} \right \|}{{\left \| {\boldsymbol \omega} \right \|}^2} {\boldsymbol \omega}^{\wedge} + \frac{\left \| {\boldsymbol \omega} \right \|-\sin \left \| {\boldsymbol \omega} \right \|}{{\left \| {\boldsymbol \omega} \right \|}^3} ({\boldsymbol \omega}^{\wedge})^2 \right]{\mathbf v},
  \end{equation}
}
\hspace{-0.39em}where ${\boldsymbol \omega}^{\wedge}$ denotes the skew-symmetric matrix of ${\boldsymbol \omega}$. We exploit SE(3) to describe facial shape morphs for its superior capability in handling mandibular rotations and higher robustness to pose perturbations than the common translation deformation
${\mathbf x}'={\mathbf x}+{\mathbf t}$.

\subsubsection{Hyper Nets} To obtain a more compact and expressive latent space, we introduce a meta-learning approach~\cite{sitzmann2019siren}. A Hyper Net $\phi_n$ is implemented by an MLP and it predicts the instance-specific parameters for ExpNet $\mathcal{E}$, IDNet $\mathcal{I}$ and DetailNet $\mathcal{D}$. It takes a latent code ${\mathbf z}$ as input and generates the parameters for the neurons in a Mini-Net ${\boldsymbol \psi}_n$ so that the learned facial representations possess a higher variety. 

\begin{figure*}
  \centering
  \begin{subfigure}[b]{0.33\textwidth}
    \includegraphics[width=1\linewidth]{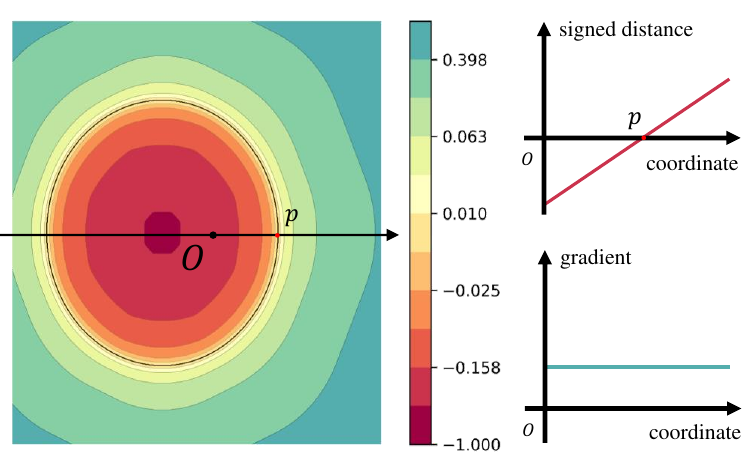}
    \caption{Signed Distance Field}
  \end{subfigure}
  \hfill
  \begin{subfigure}[b]{0.33\textwidth}
    \includegraphics[width=1\linewidth]{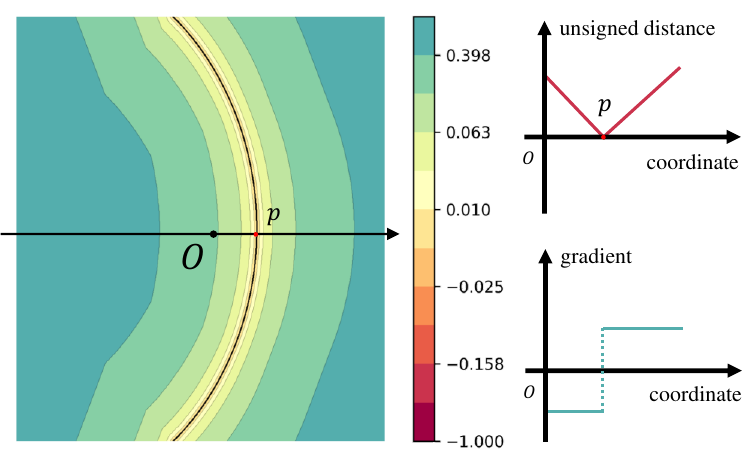}
    \caption{Unsigned Distance Field}
  \end{subfigure}
  \hfill
  \begin{subfigure}[b]{0.33\textwidth}
    \includegraphics[width=1\linewidth]{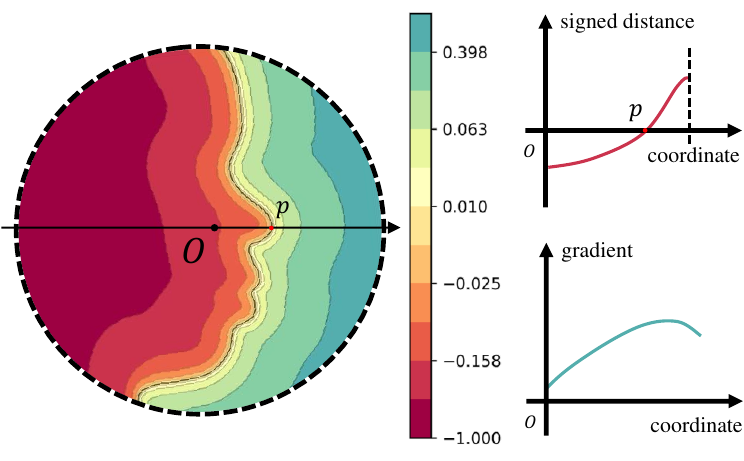}
    \caption{SDF on pseudo watertight face}
  \end{subfigure}

  \caption{\textbf{(a)} SDF is able to represent closed shapes. \textbf{(b)} UDF is capable of representing an open surface but the gradient is discontinuous at the boundary, making it hard to be fitted by neural networks. \textbf{(c)} The proposed method generates pseudo watertight faces and restricts implicit functions on them, enabling implicit neural networks to learn geometry representations on 3D faces.}
  \label{fig:wt}
\end{figure*}

\subsection{Improved Expression Embedding Learning}
\label{subsec:emb}
The auto-decoder framework proposed by \cite{park2019deepsdf} has been widely adopted in INRs to jointly learn embeddings and network parameters. In the previous attempt \cite{yenamandra2021i3dmm}, each expression type is encoded by one embedding for attribute disentangling. Unfortunately, such an embedding is merely able to represent the average shape morph of an expression type, making the learned latent space fail to capture more diverse deformation details across individuals.
To avoid the dilemma above, we improve the learning strategy by treating each face scan as a unique expression and generating a specific embedding for it. In this way, the latent space is significantly extended, which enables $\mathcal{E}$ to represent more fine-grained deformations. On the other side, there exists a potential failure mode that the identity properties are tangled into the expression space again, and $\mathcal{I}$ collapses to an identity mapping. To tackle this challenge, we suppress $\mathcal{E}$ when the current training sample is a neutral face, written as:
\begin{equation}
  \mathcal{E}({\mathbf p_{ne}}, {\mathbf z}_{exp}, l)\equiv{\mathbf p}_{ne},
\end{equation}
where ${\mathbf p_{ne}}$ denotes a point from a neutral face. By applying such a learning strategy, $\mathcal{I}$ and $\mathcal{T}$ jointly learn shape representations on neutral faces, and $\mathcal{E}$ focuses only on expression deformations. Moreover, only neutral labels are required during training, bypassing the need for dense expression labels.
\subsection{Loss Functions}
ImFace++ is trained with a set of loss functions to learn plausible facial shape representations, dense correspondence, and high-frequency details.

\textbf{Reconstruction Loss.} The basic SDF structure loss is applied to learn implicit fields:
{\small
  \begin{equation}
    {\cal L}_{sdf}^i= 
    {\lambda_1} \sum_{{\mathbf p} \in \Omega_i} \left| f({\mathbf p})-{\bar s} \right|
    + {\lambda_2} \sum_{{\mathbf p} \in \Omega_i} (1-\left \langle {\nabla f({\mathbf p})} , {\mathbf{\bar n}} \right \rangle),
  \end{equation}
}
\hspace{-0.38em}where ${\bar s}$ and ${\mathbf{\bar n}}$ denote the ground-truth SDF values and the field gradients, respectively. $i$ denotes the index of sampled faces and $\lambda$ indicates the trade-off parameter. ${\Omega_i}$ is the sampling space of the scan $i$.

\textbf{Eikonal Loss.} To obtain reasonable fields throughout the network, multiple Eikonal losses are used to enforce the L-2 norm of spatial gradients to be unit:
{\small
  \begin{equation}
    \begin{aligned} 
      {\cal L}_{eik}^i \! = &
      {\lambda_3} \! \sum_{{\mathbf p} \in \Omega_i} \! \big(
        | \| \nabla f({\mathbf p})\| \!-\!1 |
        \!+\! | \| \nabla \mathcal{T}(\mathcal{I}({\mathbf p'}))\| \!-\!1 | 
      \big),
    \end{aligned}
  \end{equation}
}

\noindent where ${\cal L}_{eik}^i$ serves the purpose of enabling the network to satisfy the Eikonal constraint\cite{icml2020_2086} in both the expression and identity spaces concurrently. This constraint ensures that both the spaces adhere to the SDF properties, where the gradient norm is uniformly unit, thereby preserving the physical integrity of the deformation field.

\textbf{Embedding Loss.} It regularizes the embedding with a zero-mean Gaussian prior: 
{\small
\begin{equation}
    {\cal L}_{emb}^i = 
    {\lambda_4} \left( {\|{\mathbf z}_{exp}\|}^2+{\|{\mathbf z}_{id}\|}^2 \right) + {\lambda_5}{\|{\mathbf z}_{detail}\|}^2.
\end{equation}}

\textbf{Landmark Generation Loss.} The ${\mathit l}_1$-loss is used to learn the Landmark-Nets $\eta$, $\eta'$:
{\small
\begin{equation}
  {\cal L}_{lmk_g}^i\!=\!{\lambda_6} 
  \sum_{n=1}^k \left( |l_n-{\bar {l_n^i}}|+|l'_n-{\bar {l_n'}}| \right),
\end{equation}
}
\hspace{-0.39em}where ${\bar {l^i}}$ denotes the $k$ labeled landmarks on sample $i$, ${\bar {l'}}$ denotes the landmarks on the corresponding neutral face.

\textbf{Landmark Consistency Loss.} We exploit this loss to guide the deformed landmarks to be located at the corresponding positions on the ground-truth neutral and template faces for better correspondence performance: 
{\small
\begin{equation}
  {\cal L}_{lmk_c}^i\!=\!{\lambda_7} 
  \sum_{n=1}^{m} \left( |\mathcal{E}(l_n)-{\bar {l'}}_n|+|\mathcal{I}(\mathcal{E}(l_n))- {l_n''}| \right),
\end{equation}
}
\hspace{-0.39em}where $m$ denotes the point number of each batch sampled.

\textbf{Residual Constraint.}
As in \cite{deng2021deformed}, we apply a penalty to $\delta$ to prevent it from excessively acquiring template face information, which could potentially compromise the quality of the morphable model. This penalty is implemented by:
{\small
  \begin{equation}
    {\cal L}_{res}^i= 
    {\lambda_8} \sum_{{\mathbf p} \in \Omega_i} |\delta({\mathbf p})|.
  \end{equation}
}

\textbf{Improved Expression Embedding Constraint.}
As described in Sec.~\ref{subsec:emb}, to avoid a potential failure mode that the identity properties are tangled into the expression space, we explicitly disentangle ExpNet $\mathcal{E}$ and IDNet $\mathcal{I}$ by:
{\small
  $${\cal L}_{imp}^i = {\lambda_9} \sum_{{\mathbf p} \in \Omega_i} \begin{cases}
||\mathcal{E}({\mathbf p}) - {\mathbf p}||^{2}, & {if\;i\;is\;a\;neutral\;face}\\
0. & else \\
\end{cases}$$
}

In the initial training phase, our focus lies on training the disentangled deformation fields, as elaborated in Sec.~\ref{subsec:disentangled}. This phase involves learning the coarse base SDF and establishing correspondences between various facial shapes. The training loss is formulated as follows:
{\small
  \begin{equation}
    {\cal L}_{\hat{f}} = \sum_i ({\cal L}_{sdf}^i+{\cal L}_{eik}^i+{\cal L}_{emb}^i+{\cal L}_{lmk_g}^i+{\cal L}_{lmk_c}^i+{\cal L}_{res}^i+{\cal L}_{imp}^i).
  \end{equation}
}

Subsequently, we freeze the sub-networks $\mathcal{E}$, $\mathcal{I}$, and $\mathcal{T}$ and proceed to train $\mathcal{D}$ to learn the RDF, by substituting $\hat{f}$ with $f$ in the following training loss: 
{\small
  \begin{equation}
    {\cal L}_{f} = \sum_i ({\cal L}_{sdf}^i+{\cal L}_{eik}^i+{\cal L}_{emb}^i).
  \end{equation}
}

In alignment with the progressive training strategy in~\cite{yifan2021geometry}, we combine ${\cal L}_{f}$ and ${\cal L}_{\hat{f}}$ via $\kappa {\cal L}_{f}+ (1-\kappa){\cal L}_{\hat{f}}$ using a weighted sum $\kappa=\frac{1}{2}(1+\cos(\pi \frac{t-T_{m}}{1-T_{m}}))$, where $T_{m}\in [0, 1]$ and $t\in [T_{m}, 1]$ indicate the training percentile and the current training progress, respectively. We symmetrically adjust the learning rates, decreasing or increasing them in proportion to the loss weight.

At the testing phase, for each 3D face indexed by $j$, we minimize the following objective to obtain its latent embeddings and the reconstructed 3D face:
{\small
  \begin{equation}
    \mathop{\arg\min}\limits_{{\mathbf z}_{exp},{\mathbf z}_{id},{\mathbf z}_{detail}} \sum_j ({\cal L}_{sdf}^j+{\cal L}_{eik}^j+{\cal L}_{emb}^j).
  \label{fit}
  \end{equation}
}

\section{Experiments}
We conduct thorough subjective and objective evaluation of ImFace++, along with ablation studies to validate the effectiveness of the specifically designed modules. Finally, we demonstrate various applications of ImFace++.

\subsection{Dataset}

FaceScape~\cite{yang2020facescape} is used for evaluation. It is a large-scale high quality 3D face dataset consisting of 938 individuals with 20 types of expressions. The data from 847 individuals are publicly accessible and we mainly adopt them for experiments. More specifically, 16,434 face scans from 831 individuals are sampled as the training set, and 319 face scans from the remaining 16 individuals are utilized as the testing set, where all the types of expressions are almost included. It is worth noting that, before conducting any data preprocessing, we make sure that the face scans include high-frequency details by making use of the displacement map provided by FaceScape. Additionally, 100 face scans from the LYHM~\cite{dai2020statistical} dataset are employed for cross-dataset evaluation.

\begin{figure}
  \centering
  \setlength{\abovecaptionskip}{0pt}
  \setlength{\belowcaptionskip}{0pt}
  \includegraphics[width=1.0\linewidth]{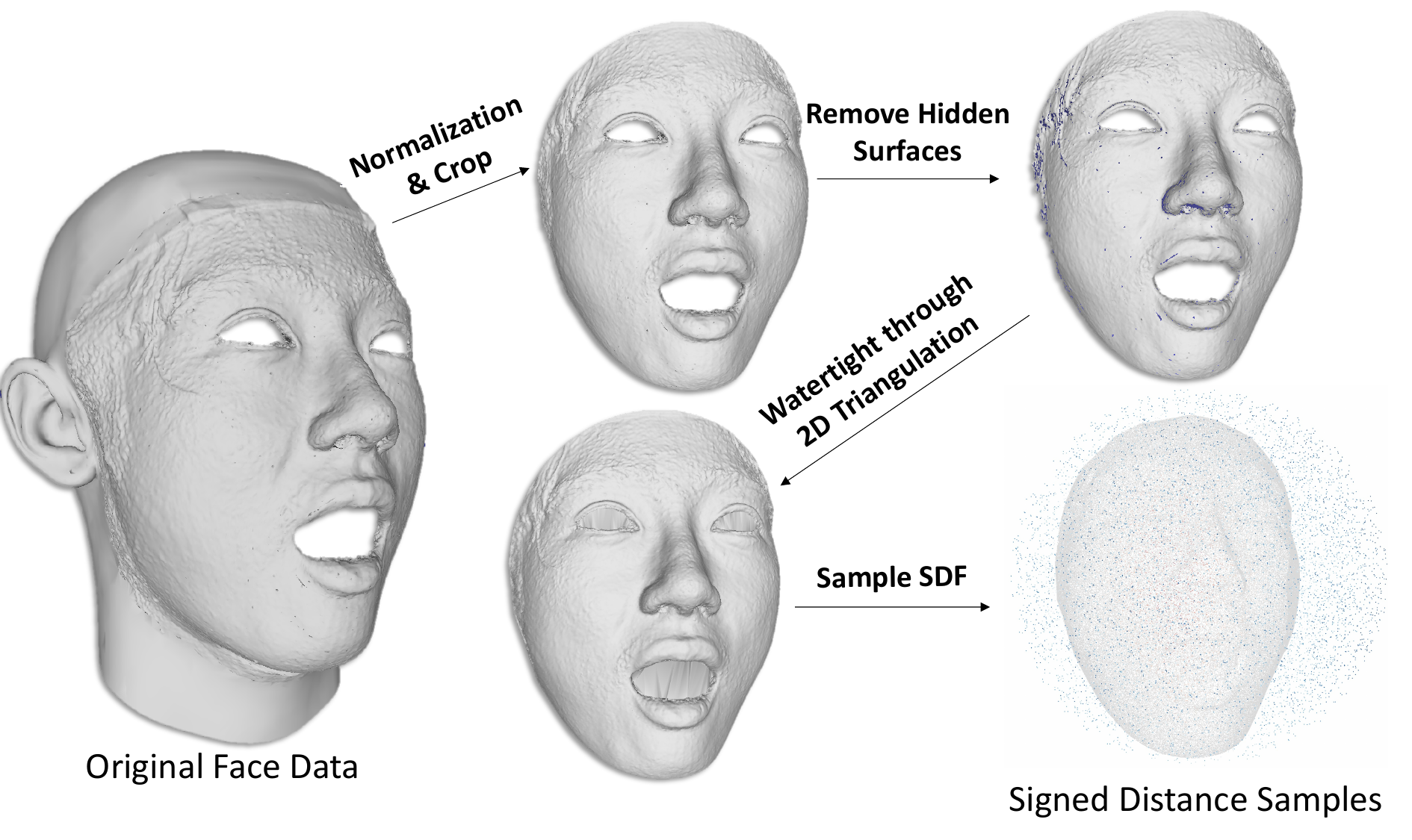}
   \caption{Illustration of the preprocessing pipeline. The original face data from FaceScape~\cite{yang2020facescape} undergo initial normalization and cropping. Hidden surfaces are subsequently removed. The Delaunay Triangulation algorithm~\cite{lee1980two} is then employed to achieve watertight results. Lastly, we sample points and estimate the SDF and normals for training ImFace++.}
   \label{fig:preprocess}
\end{figure}

\textbf{Data Preprocessing.} Given that neural networks are particularly adept at fitting functions that are differentiable across the entire domain, recent studies on implicit functions generally require watertight input data. While functions like UDF do not strictly necessitate watertightness, they exhibit a non-differentiability when crossing a surface and may struggle to handle fine details, as illustrated in Fig.~\ref{fig:wt}. To address this, we develop an efficient preprocessing pipeline that transforms the input data into pseudo-watertight faces. This preprocessing step enables the establishment of a general SDF on these faces. By doing so, we ensure that the geometry and correspondence are learned with the same level of precision and fidelity as that of watertight objects.

As shown in Fig.~\ref{fig:preprocess}, the preprocessing pipeline consists of a series of steps: i) We rigidly align the facial scans to a frontal orientation using facial landmarks. Each mesh is then normalized to a standard unit of 10 cm. ii) The coordinate origin is defined at a point 4 cm behind the nose tip. A sphere with a radius of 10 cm is set as the sampling area. Mesh triangles outside of this sphere are cropped. iii) We apply the Ray-Triangle Intersection algorithm~\cite{moller1997fast} to remove hidden surfaces, such as the nasal and oral cavities. iv) Delaunay Triangulation\cite{lee1980two} is performed on the x-y coordinates of the mesh to create an oriented and pseudo-watertight structure. v) Since FaceScape provides 3D face scans with a uniform topology, we sample 13,008 points as landmarks within the defined sampling area of each scan to train ImFace++.

With the pseudo-watertight facial structures generated, we compute the SDF values through a distance transform. The sign of each sample is determined by the angle between its distance vector to the nearest surface and the positive z-axis direction. The values behind the facial surface are defined as negative. To ensure comprehensive coverage, we uniformly sample 250,000 points on each facial surface: 125,000 points near the facial surface with random displacements sampled from a normal distribution $\mathcal{N}(0, 1 cm)$, and 125,000 points within the sphere. For each of these points, we calculate their signed distances and gradient vectors. In the end, the sampled data are organized into triplets, including the query point, gradient vectors, and signed distance value. These triplets serve as the input data for training ImFace++.

\subsection{Implementation Details}

\textbf{Network Architecture.} All Mini-Net blocks used by ExpNet, IDNet, and TempNet, are implemented as Multi-Layer Perceptrons (MLPs) with 3 hidden layers, each consisting of 128-dimensional hidden features, while the Mini-Net block utilized by DetailNet is implemented as MLPs with 3 hidden layers, comprising 256-dimensional hidden features. For the activation function, we employ sine activation in every Mini-Net, and the parameters are initialized as described in \cite{sitzmann2019siren}.
The frequency hyperparameter $w$ in the sine activation function determines the capability of the neural network. We use a smaller value of $w=30$ for the disentangled deformation fields and a larger value of $w=60$ for DetailNet to enhance the capability of the networks. At the end of each Mini-Nets block, the Fusion Network is implemented as a 128-dimensional, 3-layer MLP with a softmax activation function. During training, the parameters of Mini-Nets are generated by corresponding Hyper Nets. This is adopted to achieve a more expressive latent space, following the insights from \cite{deng2021deformed} and \cite{sitzmann2019metasdf}. The Hyper Nets consist of 1-layer MLPs activated by the Rectified Linear Unit (ReLU) function, with the hidden layer dimensionality set to 32. The Landmark-Nets are composed of three 256-dimensional fully connected layers. The identity, expression, and detail embeddings are individually represented as 160-dimensional vectors.

\textbf{Training and Inference Settings.} The trade-off parameters $\lambda_1,\cdots,\lambda_9$ to train the networks are set to $3e3$, $1e2$, $5e1$, $1e5$, $1e3$, $1e3$, $1e2$, $1e2$, and $1e4$, respectively. The model is optimized with Adam in an end-to-end manner. We train the model for 1,500 epochs with a learning rate of 0.0001. The training phase takes around 4 days on 4 32G NVIDIA V100 GPUs with minibatches of size 48. We randomly sample $m=4096$ points from the total 13,008 points in each batch to train the landmark consistency loss. During testing, it takes about 8 hours to infer 319 samples on a single GPU. We use Marching Cubes~\cite{lewiner2003efficient} to reconstruct facial surfaces from the signed distance field, where the voxel resolution is set to $256^3$. All the meshes are rendered by Pyrender\cite{pyrender}.

\begin{figure*}
  \centering
  \includegraphics[width=1\linewidth]{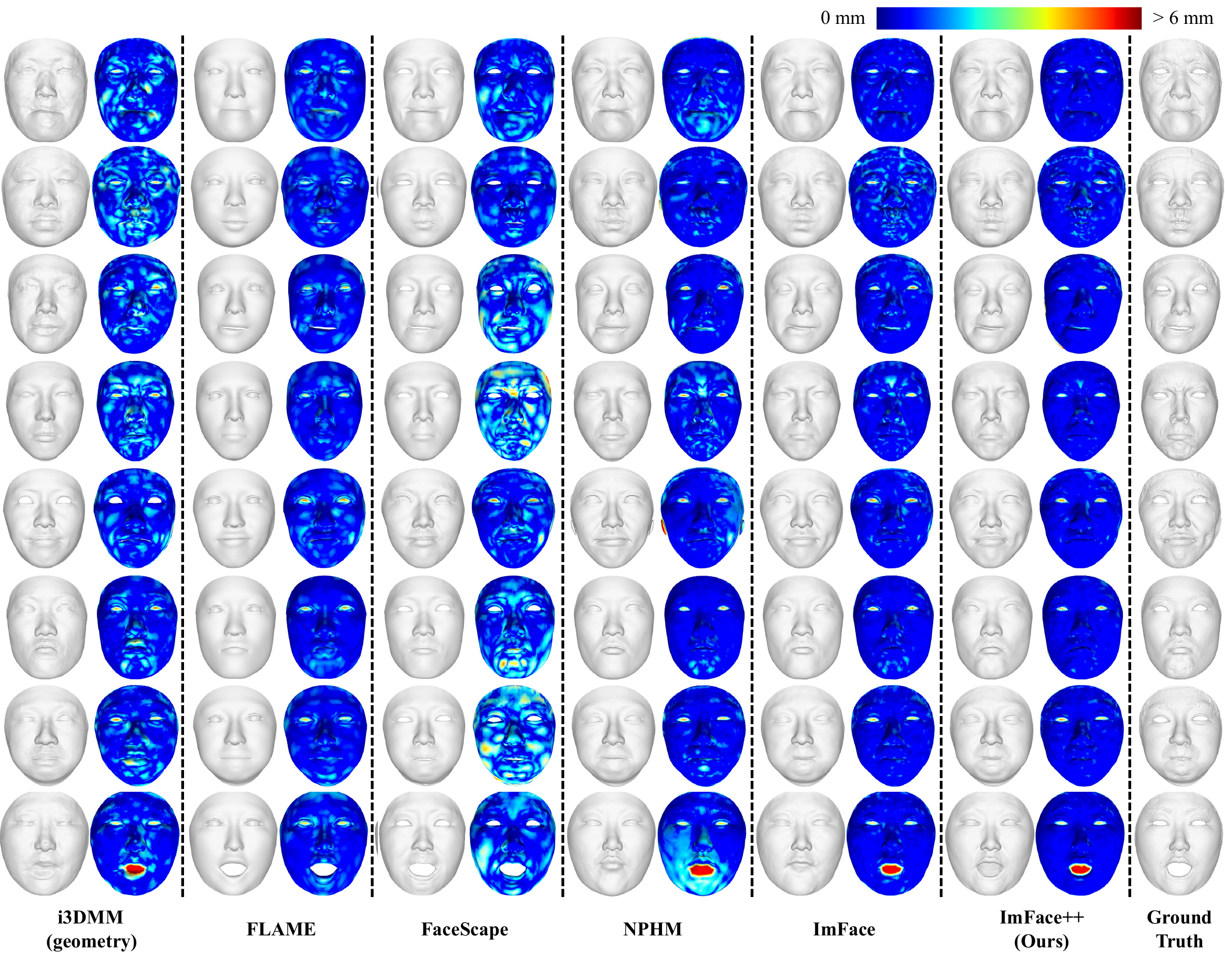}
   \vspace{-8mm}
  \caption{Reconstruction comparison with i3DMM\cite{yenamandra2021i3dmm}, FLAME\cite{li2017flame}, FaceScape\cite{yang2020facescape}, NPHM\cite{giebenhain2022learning}, and ImFace\cite{zheng2022imface}. Each column corresponds to a person with a non-neutral expression. Visually inspected, ImFace++ captures much richer shape variations and high-frequency details with more compact latent embeddings. Zoom in for a better view.}
  \label{fig:experiment_recon}
\end{figure*}

\subsection{Comparison with State-of-the-Art}
\subsubsection{Reconstruction}

We employ the ImFace++ model to fit face scans by optimizing Eq.~(\ref{fit}) and then compare the reconstruction results with several existing methods, including FLAME~\cite{li2017flame}, FaceScape~\cite{yang2020facescape}, the geometry model of i3DMM~\cite{yenamandra2021i3dmm}, NPHM~\cite{giebenhain2022learning}, and our previous version, ImFace~\cite{zheng2022imface}, which represent the current state-of-the-art. The official code of FLAME is used to fit the full face scans in the test set, utilizing 300 identity parameters and 100 expression parameters. For FaceScape, we use their released bilinear model built from 938 individuals for testing. The model parameters consist of 300 identity parameters and 52 expression parameters. It is important to note that our test scans are part of the training set for FaceScape. Furthermore, we modify their official code to fit full scans, rather than just landmarks, for improved results. For i3DMM and ImFace, we retrain these models on the same training set as used for ImFace++ to ensure fair comparison. In both cases, the identity and expression embeddings are set to 160 dimensions, which aligns with our settings. For NPHM, we utilize the model trained with the official code. To maintain the consistency with our study where we focus only on the facial region, 22 facial anchors are utilized and the parameter $K_{symm}$ in NPHM is set to 8.

\textbf{Qualitative Evaluation.} Fig.~\ref{fig:experiment_recon} provides a visual demonstration of the reconstruction results obtained by various models, with each row corresponding to a test individual displaying a non-neutral expression. i3DMM, the first deep implicit model for human heads, exhibits limitations in capturing complex deformations and high-frequency details when confronting with relatively intricate scenarios involving unseen facial shapes, incurring artifacts on the reconstructed results. FLAME and FaceScape, while effective in certain aspects, are constrained by fixed resolutions and uniform mesh topologies. FLAME excels at representing identity characteristics but struggles with non-linear deformations, leading to rigid facial expressions. FaceScape delivers more favourably results, primarily owing to high-quality training scans and the inclusion of test faces in the training set. However, it still fails to accurately portray expression morphs and high-frequency details.
NPHM utilizes local fields to model the SDF in a neutral expression, but the globally conditioned deformation field makes it challenging to represent fine-grained deformations and high-frequency details.
Comparatively, ImFace excels in reconstructing facial shape by using the disentangled deformation fields, preserving subtle and rich nonlinear facial muscle deformations with fewer identity and expression latent parameters, such as frowns and pouts. However, it falls short of representing high-frequency details.
In contrast, ImFace++ significantly improves upon its predecessor. The introduction of RDF within the template space enables it to effectively handle intricate features, as demonstrated by its success in representing details such as eyelids (visible in the first two rows in Fig.~\ref{fig:experiment_recon}) and wrinkles (visible in the third to fifth rows).

\begin{figure}
  \centering
  \setlength{\abovecaptionskip}{0pt}
  \setlength{\belowcaptionskip}{0pt}
  \includegraphics[width=1.0\linewidth]{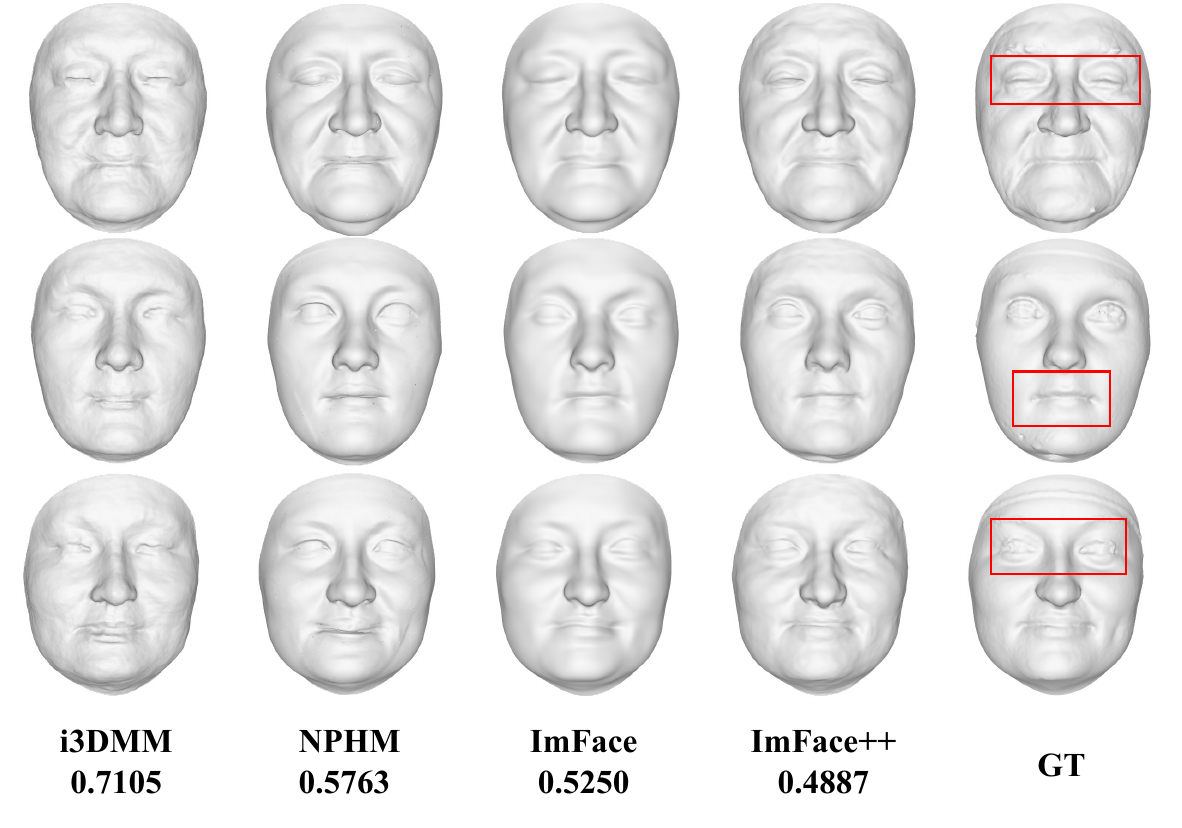}
   \caption{Cross-dataset reconstruction results. The model is trained on FaceScape and tested on LYHM. The Chamfer distance for each method is listed below.}
   \label{fig:LYHM}
\end{figure}

Additionally, we select 100 facial shapes from the LYHM dataset for cross-dateset evaluation. We calculate the rotation, translation, and scale for each shape using 3D landmarks. We then register these facial shapes to our coordinate system and preprocess them as described in Sec.~4.1. The evaluation is performed using the model trained on the FaceScape dataset. We present the cross-dataset results in Fig.~\ref{fig:LYHM}, demonstrating that our model generalizes well to new data.

\begin{table}
  \centering
  \setlength{\belowcaptionskip}{0pt}
  \begin{tabular}{@{}ccccc@{}}
    \toprule
    Metrics & Dim. & {Chamfer($mm$)$^{\dag}$} &  {F-score@1$mm$$^{\P}$} & N.C.$^{\P}$ \\
    \midrule
    i3DMM\cite{yenamandra2021i3dmm} & 320 & 0.759 & 83.87 & 0.938 \\
    FLAME\cite{li2017flame} & 400 & 0.662 & 88.18 & 0.875 \\
    FaceScape\cite{yang2020facescape} & 352 & 0.731 & 79.38 & 0.908\\
    NPHM\cite{giebenhain2022learning} & 1000 & 0.637 & 91.67 & 0.934 \\
    ImFace\cite{zheng2022imface} & 320 & 0.553 & 94.67 & 0.950\\
    ImFace++ & 480 & \bf{0.511} & \bf{96.11} & \bf{0.956}\\
    \bottomrule
  \end{tabular}
  \caption{Quantitative reconstruction comparison with the state-of-the-art methods ($^{\dag}$Lower is better; $^{\P}$Higher is better).}
  \label{table:quantitative}
\end{table}

\textbf{Quantitative Evaluation.} To ensure fair comparison, all faces in this evaluation undergo preprocessing to remove inner structures like eyeballs. This step allows for the computation of quantitative metrics within the same facial region for all models. Specifically, the symmetric Chamfer distance, F-score, and normal consistency (N.C.) are employed as metrics, with a strict F-score threshold set at 1mm. The results are presented in Tab.~\ref{table:quantitative}, clearly demonstrating that ImFace++ outperforms its counterparts by a significant margin across all the metrics, unequivocally validating its effectiveness. Additionally, a color-coded distance (fit-to-scan) is used to visualize the reconstruction errors in Fig.~\ref{fig:experiment_recon}. Overall, ImFace++ achieves more accurate face reconstruction results compared to other methods, particularly excelling in capturing nonlinear deformations and high-frequency facial details.

\subsubsection{Correspondence}
\label{correspond}
In contrast to existing methods that typically rely on accurate face registration, ImFace++ leverages the disentangled deformation fields to automatically learn correspondences among different face scans. Furthermore, we incorporate the training critic, \ie the landmark consistency loss ${\cal L}_{lmk_c}^i$, to enhance the feature. This evaluation is designed to assess the correspondence establishment capabilities of various INR-based models.
In particular, given two 3D faces, we use i3DMM, ImFace, and ImFace++ to fit them, subsequently deforming the densely sampled points to the template space. This process ensures point-to-point correspondences, which are achieved through a nearest neighbor search using the KD-Tree algorithm for efficiency. Considering that NPHM is a forward-deformation-based method, we deform the sampled points in a neutral expression to another expression and examine the correspondences.

\begin{figure*}
  \centering
   \includegraphics[width=1\linewidth]{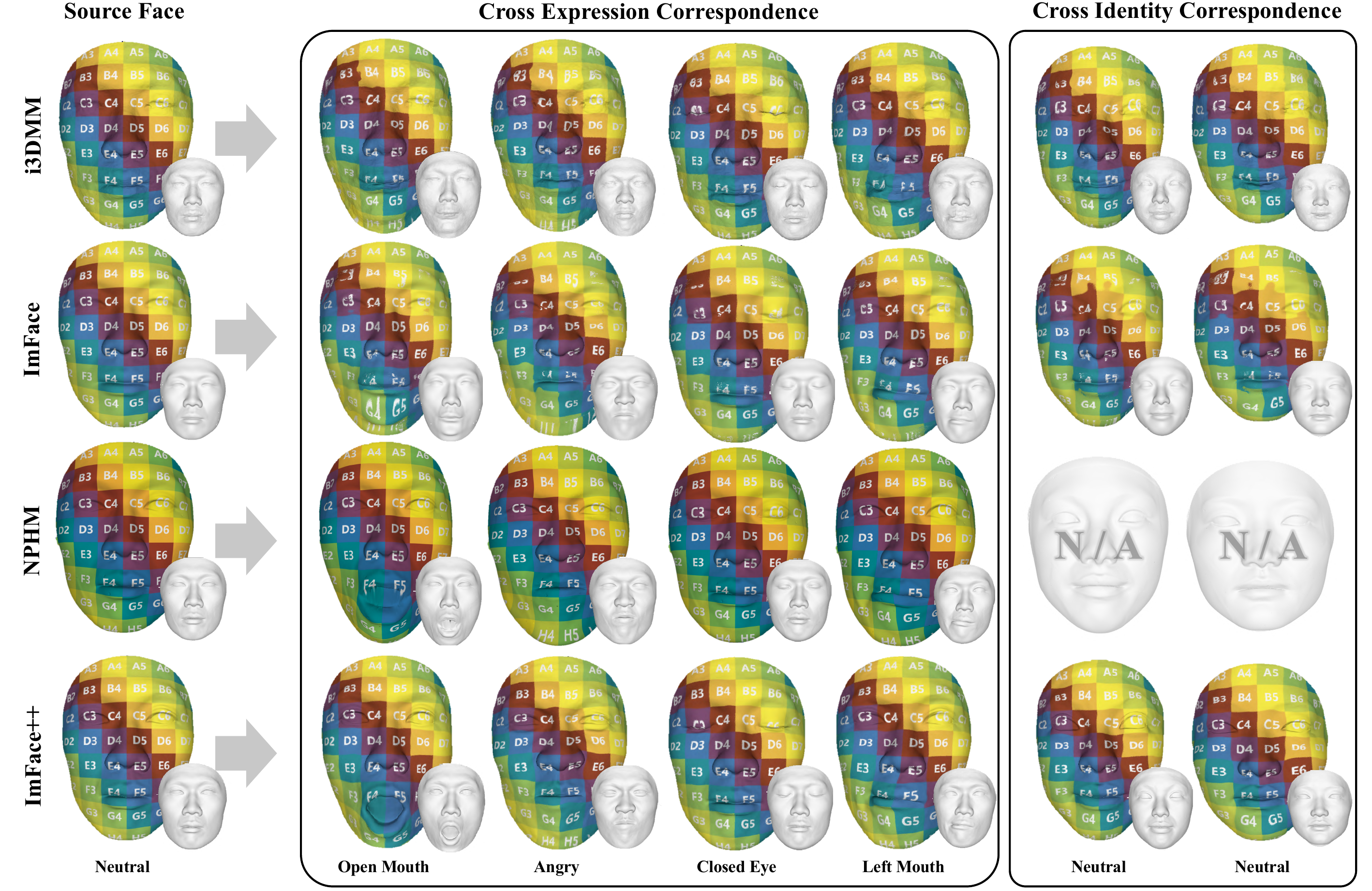}
    \vspace{-4mm}
   \caption{Comparison of correspondences with i3DMM\cite{yenamandra2021i3dmm}, ImFace\cite{zheng2022imface}, and NPHM~\cite{giebenhain2022learning}. The leftmost face is morphed into multiple expressions (middle columns) and different identities (rightmost columns). ImFace++ surpasses the other backward-deformation methods, \ie i3DMM and ImFace, and achieves the results comparable to the forward-deformation method NPHM. Due to the forward-deformation mechanism, there are no cross-identity correspondence results for NPHM. Zoom in for a better view.}
   \label{fig:corr}
\end{figure*}

\textbf{Qualitative Evaluation.} Fig.~\ref{fig:corr} provides visual comparison of the results, with manually painted color patterns on the shapes to facilitate quality assessment. This is achieved by mapping the colors from a source face to the target one utilizing the established point-to-point correspondence. In alignment with our approach, i3DMM incorporates a landmark consistency loss to enhance correspondences. However, it employs a single deformation field to represent both expression and identity deformations in an intertwined manner. Consequently, the correspondences may not be accurately learned in certain areas, such as the mouth (regions F4-F5 in the painted color patterns), forehead (regions B3-B6), and nostril (regions E4-E5).
ImFace successfully establishes correspondences across various expressions and identities. Nonetheless, it is noticeable that minor internal texture discrepancies occasionally occur around the mouth (regions F4-F5, G4-G5). This is primarily because facial shapes undergo significant changes in these localized areas with different expressions.
ImFace++ addresses this challenge by employing densely sampled landmarks (increasing from 68 in ImFace to 4,096 in ImFace++) to supervise the landmark consistency loss during training, resulting in more accurate outcomes.
Due to the forward-deformation pipeline, NPHM naturally exhibits more consistent correspondences than backward-deformation approaches. However, it may overlook exaggerated expressions like a closed eye, potentially leading to semantically incorrect correspondences (regions G4-G5). In contrast, ImFace++ achieves comparable or even superior results to NPHM, demonstrating its effectiveness.
\begin{table}
  \centering
  \setlength{\belowcaptionskip}{0pt}
  \begin{tabular}{@{}ccc@{}}
    \toprule
    Metrics & {EDE ($mm$)$^{\dag}$} & {TDE ($mm$)$^{\dag}$} \\
    \midrule
    i3DMM\cite{yenamandra2021i3dmm} & - & 4.474  \\
    NPHM\cite{giebenhain2022learning} & 3.130 & -  \\
    ImFace\cite{zheng2022imface} & 2.805 & 3.116 \\
    ImFace++ & \bf{2.383} & \bf{1.620} \\
    \bottomrule
  \end{tabular}
  \caption{Quantitative correspendence comparison with the state-of-the-art methods. ($^{\dag}$Lower is better.)}
  \label{table:corr_quantitative}
  \vspace{-2mm}
\end{table}

\begin{figure*}
  \centering
  \setlength{\abovecaptionskip}{0pt}
  \setlength{\belowcaptionskip}{0pt}
  \includegraphics[width=1.0\linewidth]{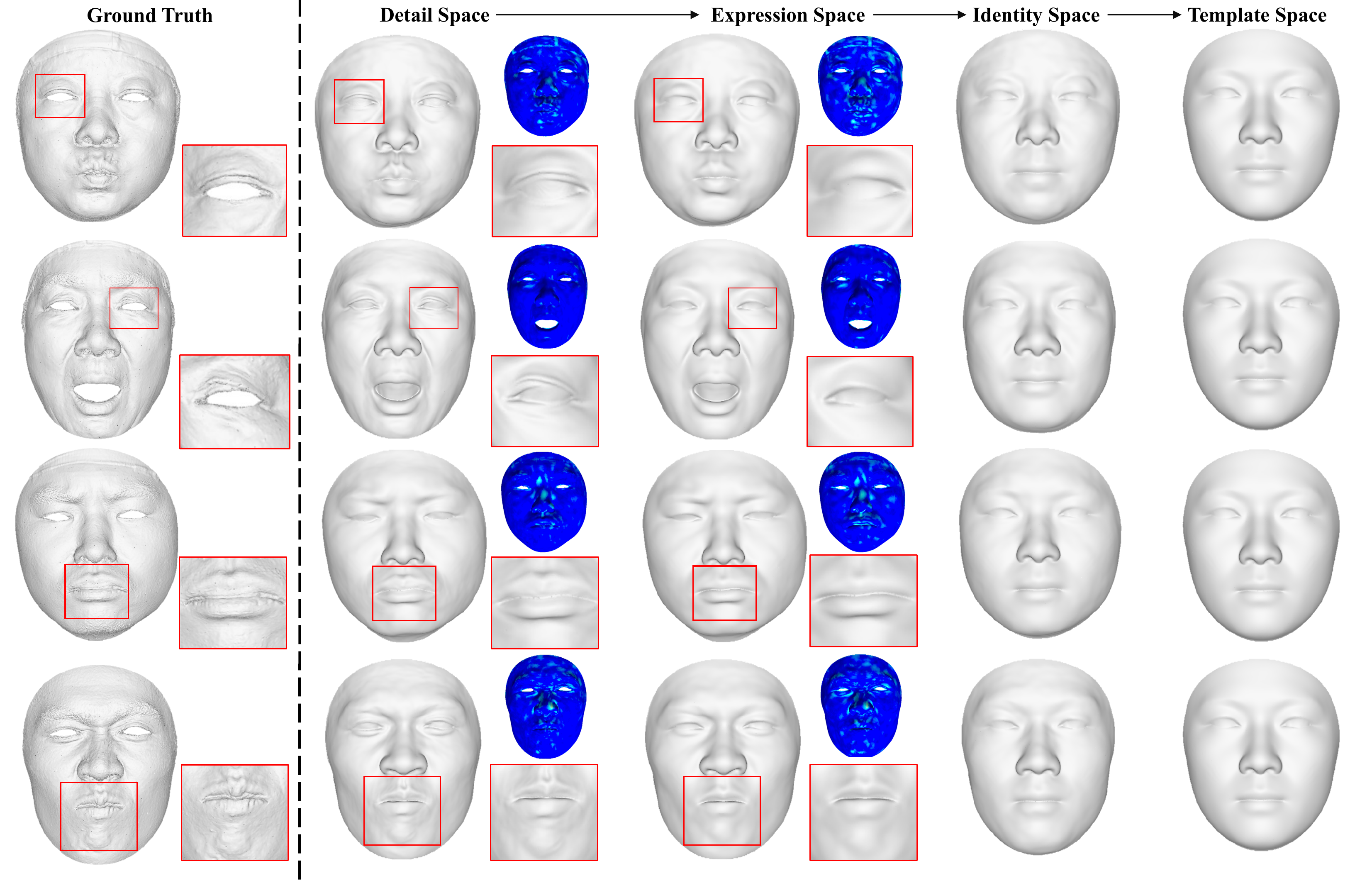}
   \caption{Module analysis. The disentangled deformation fields are responsible for learning deformations related to facial expressions and identities. The refinement displacement field enables the representation of high-frequency details.}
   \label{fig:model_analysis}
\end{figure*}

\textbf{Quantitative Evaluation.} To further validate the effectiveness of our method, we conduct quantitative evaluation on the learned correspondences similar to i3DMM~\cite{yenamandra2021i3dmm}. For each test face $F$ with identity $i$ from FaceScape, we first identify the face $F_{ne}$ with the same identity but in a neutral expression. We then upsample the registered mesh resolution and sample 22,481 points from faces $F$, $F_{ne}$, and the template face, which are not used during training, serving as the ground-truth annotations for correspondences. 
We use i3DMM, ImFace, and ImFace++ to deform the points sampled from $F$ to the identity space and the template space, respectively. We calculate the average Expression Deformation Error (EDE), which quantifies the errors between the deformed points and the ground-truth points in the identity space, and the Total Deformation Error (TDE), which measures the errors in the template space. Note that i3DMM employs a single deformation field to represent both expression and identity deformations, allowing it to compute only the TDE metric. On the other hand, NPHM models only the deformation of expression and we can calculate only the EDE metric. For NPHM, we follow the approach of SNARF~\cite{Chen_2021_ICCV} and utilize iterative root finding to locate the canonical positions of the points sampled from $F$, allowing us to compute the EDE metric.
As shown in Tab.~\ref{table:corr_quantitative}, ImFace++ outperforms the other methods in terms of correspondence quality. Notably, it even surpasses the forward-deformation method, NPHM, underscoring the effectiveness of the disentangled deformation fields, which play a crucial role in addressing the challenges associated with learning the RDF. It is worth mentioning that the results in Tab.~\ref{table:corr_quantitative} also encompass the reconstruction errors.

\begin{figure}
  \centering
  \setlength{\abovecaptionskip}{0pt}
  \setlength{\belowcaptionskip}{0pt}
  \includegraphics[width=1.0\linewidth]{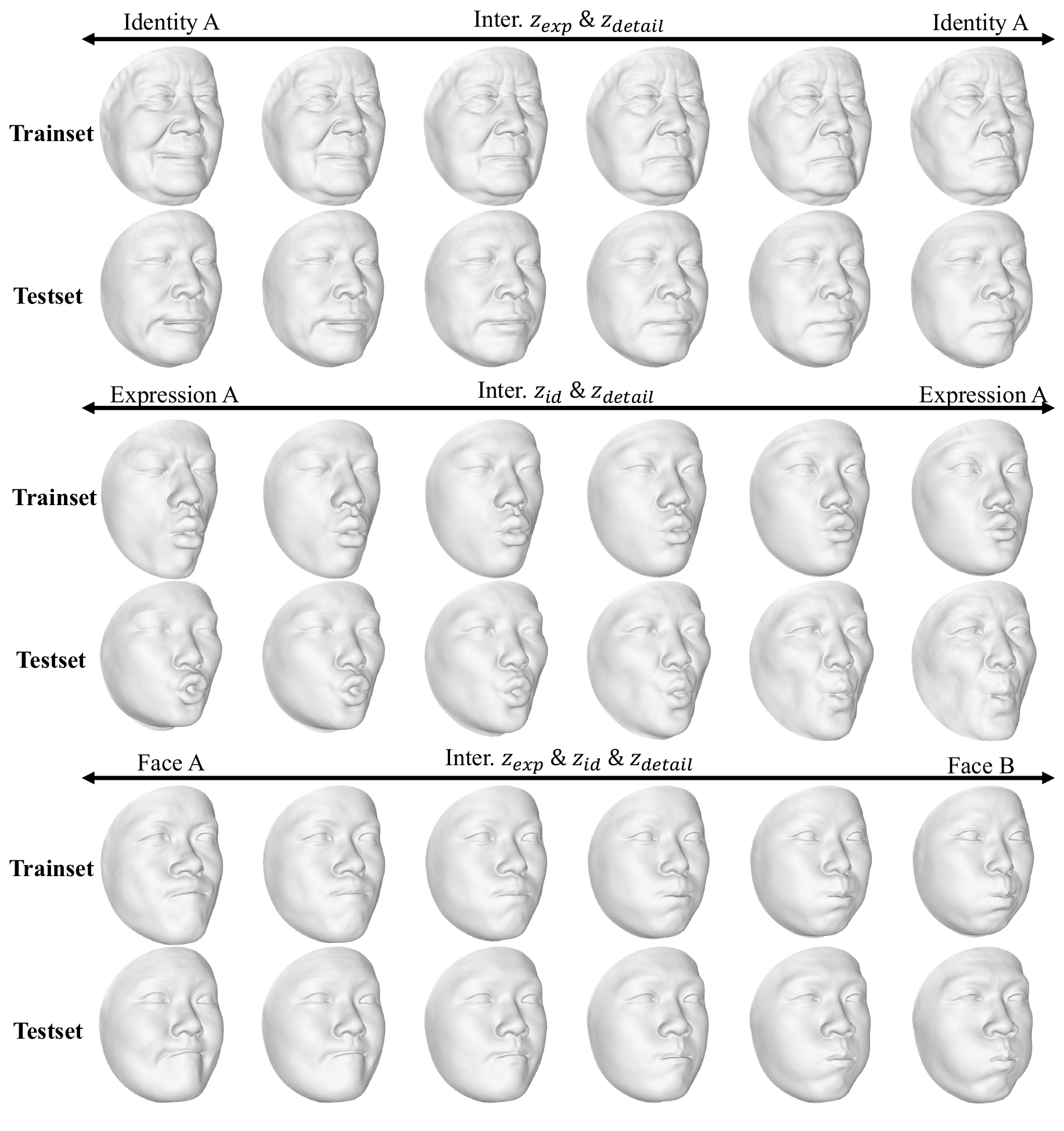}
   \caption{Embedding interpolation results. The Trainset and Testset scans can be smoothly animated by interpolating in the embedding space.}
   \label{fig:interpolation}
\end{figure}

\begin{figure}
  \centering
  \includegraphics[width=1.0\linewidth]{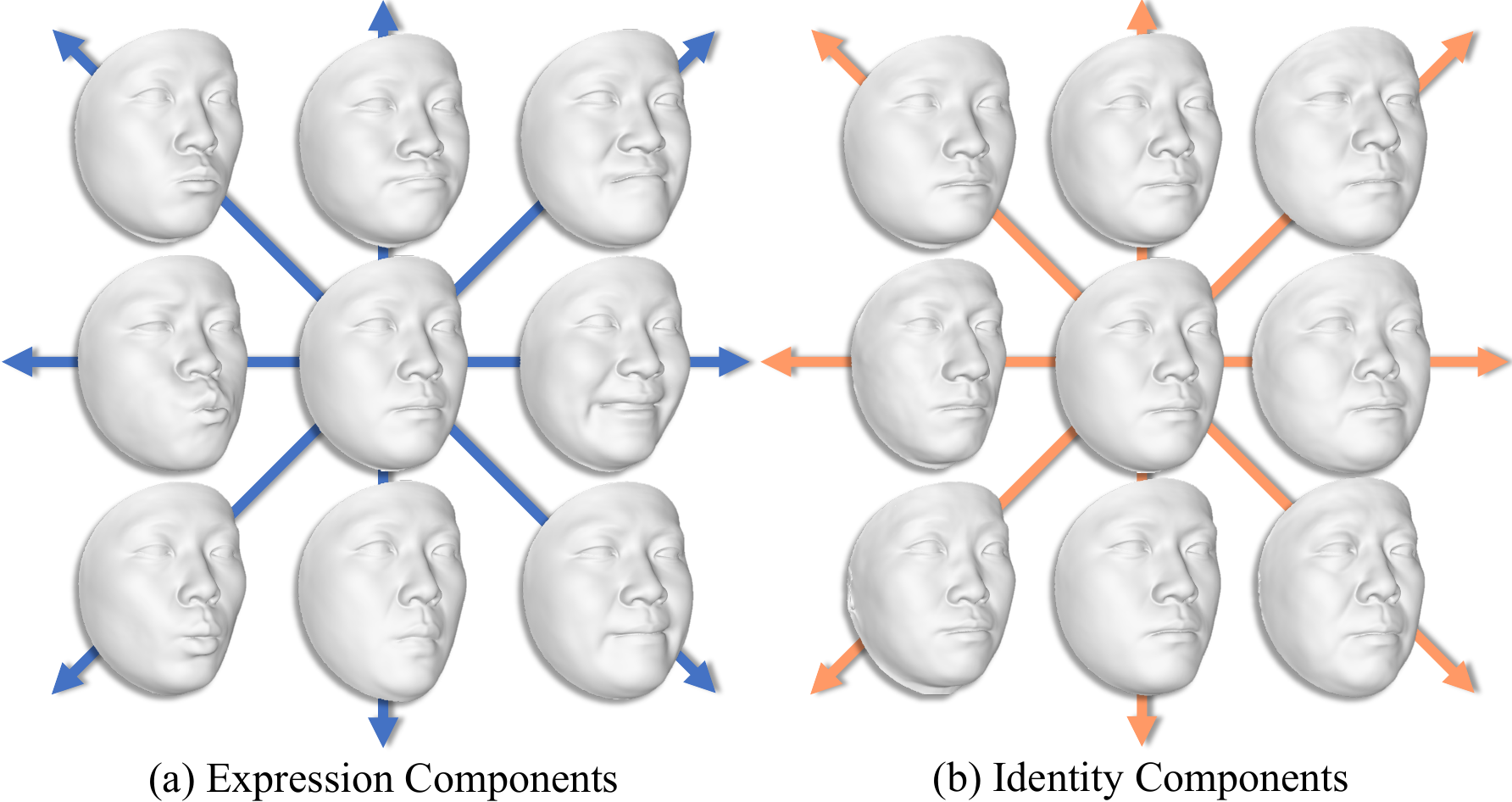}
  \caption{Visualization of the expression and identity principal components of ImFace++ on FaceScape \cite{yang2020facescape}. }
  \label{fig:pca}
\end{figure}

\begin{figure}
  \centering
  \setlength{\abovecaptionskip}{0pt}
  \setlength{\belowcaptionskip}{0pt}
  \includegraphics[width=1.0\linewidth]{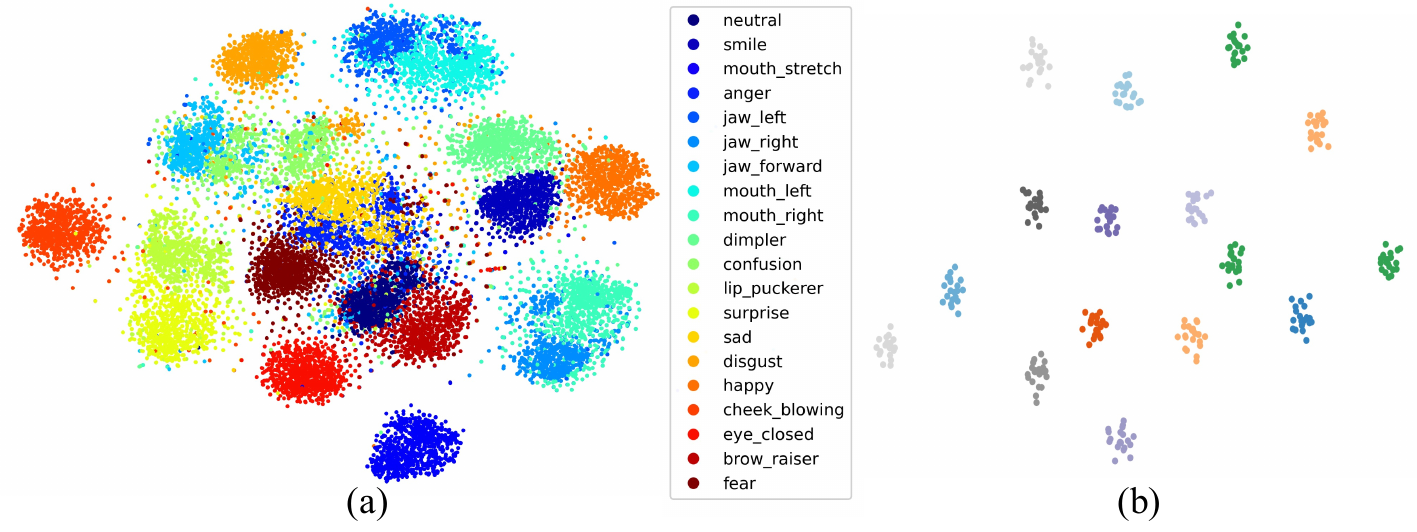}
   \caption{Visualizing the distributions of high-dimensional expression and identity embeddings using $t$-SNE. \textbf{(a)} Embedding distribution of 20 typical expressions from the training and testing sets. \textbf{(b)} The identity embedding distribution from the test set, which involves 16 individuals. }
   \label{fig:tsne}
\end{figure}

\subsection{Representation Ability}

To further assess the representation ability of ImFace++, we conduct comprehensive evaluation in both the observed and latent representation spaces. The evaluation comprises three main aspects. i) Module Analysis: we analyze the learned content of the various modules introduced in ImFace++.
ii) Compact Embedding Space Interpolation: given the compact embedding space of ImFace++, we perform animations on the generated or reconstructed facial shapes through interpolation in the identity, expression, and detail embeddings.
iii) Variation Visualization: we apply Principal Components Analysis (PCA) to the learned expression and identity embeddings to visualize model variations. Additionally, we employ $t$-SNE~\cite{Vandermaaten08a} to assess the unsupervised learning ability of ImFace++.

\textbf{Module Analysis.} Fig.~\ref{fig:model_analysis} presents the reconstructed test scans by ImFace++, offering insights into the content learned by the different modules proposed in our method. In the third and fourth columns, we can observe that the disentangled deformation fields, managed by ExpNet for expression and IDNet for identity, successfully capture rich and complex deformations. However, the reconstructed facial shapes in these cases lack high-frequency details. Nevertheless, after training the RDF within the template space, ImFace++ is capable of representing facial shapes that incorporate personalized high-frequency details. This is particularly evident in features like the eyelids in the first two rows and the mouth shape in the third row. In the ablation study section, we offer further evidences and a more comprehensive analysis.

\textbf{Compact Embedding Space Interpolation.} The seamless animation of facial shapes through interpolation in the identity, expression, and detail embeddings demonstrates that ImFace++ is able to learn compact latent embedding spaces. As illustrated in Fig.~\ref{fig:interpolation}, our identity embeddings encompass the overall shape, including the positioning of the nose, eyes, and mouth, as well as the gender of the face. In contrast, the expression embeddings capture deformations specific to facial components. Furthermore, when comparing the first and last columns in Fig.~\ref{fig:interpolation}, it becomes evident that the detail embeddings represent personalized high-frequency details of the face, such as wrinkles and dimples. As demonstrated, even when the embeddings used for interpolation exhibit significant changes over a wider range (in the last two rows), our method is capable of generating smooth and coherent results.

\textbf{Variation Visualization.} We apply PCA to the learned expression and identity embeddings to visualize the model capacity, as depicted in Fig.~\ref{fig:pca}. The standard deviations for the expression and identity embeddings are set to $\pm~0.25$ and $\pm~5.0$, respectively. The embeddings encoding facial details are fixed at the mean value calculated from the entire training set. In particular, four expression principal components are visualized in Fig.~\ref{fig:pca} (a). Despite significant expression changes, the faces maintain a consistent identity. ImFace++ is capable of producing vivid expressions by learning expression components from thousands of unique embeddings. In Fig.~\ref{fig:pca} (b), a similar phenomenon is observed in the learned identity components, where facial expressions remain stable while identity varies. These experimental results demonstrate that a good disentanglement between expression and identity is achieved, which is crucial for generating novel faces by reweighting the singular values.

Additionally, we employ $t$-SNE\cite{Vandermaaten08a} to visualize the learned high-dimensional expression embeddings from the whole training and testing scans, encompassing 20 expressions, as depicted in Fig.~\ref{fig:tsne} (a). Some entanglement remains for certain expressions, such as sadness and anger, and this is primarily due to the inherent complexity of distinguishing these similar expressions. From the perspective of facial expression recognition, these expressions often share overlapping facial features, such as furrowed brows and downturned lips, making them difficult to differentiate. Despite this, our method shows promising capabilities in differentiating between various expression types through unsupervised learning, solely from expression-related shape morphs, without the need of explicit expression labels.

\begin{table}
  \centering
  \setlength{\belowcaptionskip}{0pt}
  \begin{tabular}{@{}cccc@{}}
    \toprule
     Method & Chamfer ($mm$) $^{\dag}$ &  F-score@1$mm$$^{\P}$ & N.C.$^{\P}$  \\
    \midrule
    (a) w/o DDF & 0.568 & 94.59 & 0.950 \\
    (b) w/o Blend & 0.564 & 93.41 & 0.949 \\
    (c) w/o Extend & 0.596 & 91.75 & 0.947 \\
    (d) ImFace$^{*}$ (w/o RDF) & 0.547 & 95.33 & 0.952 \\
    (e) w/o Corres. & 0.536 & 95.37 & 0.952 \\
    (f) w/ Concat. & 0.526 & 95.52 & 0.953\\
    (g) w/ Ada. & 0.527 & 95.43 & 0.953\\
    (h) w/o ${\cal L}_{lmk_c}$ & 0.518 & 95.79 & 0.954 \\
    (i) ImFace++ & \bf{0.511} & \bf{96.11} & \bf{0.956} \\
    \bottomrule
  \end{tabular}
  \caption{Quantitative ablation study results. ImFace$^{*}$ denotes using 4,096 landmarks to supervise the landmark consistency loss during model training.}
  \label{table:ablation}
\end{table}

Furthermore, Fig.~\ref{fig:tsne} (b) presents the visualization of identity embeddings, involving 319 face scans from 16 individuals in the test set. Upon visual inspection, our model successfully captures inter-subject differences even under various complicated expressions.

\subsection{Ablation Study}
\label{ablation}
ImFace++ is constructed upon the following core components: Disentangled Deformation Fields (DDF), Refinement Displacement Field (RDF), Neural Blend-Field (Blend), Improved Expression Embedding Learning (Extend), and Hyper Net. We experimentally validate the effectiveness of these components, including evaluating the landmark consistency loss. When performing ablation of RDF, the model is essentially downgraded to ImFace. During its training process, we increase the number of the sampled points to 4,096 to supervise the landmark consistency loss, ensuring fair comparison with ImFace++.

\begin{figure}
  \centering
  \setlength{\abovecaptionskip}{0pt}
  \setlength{\belowcaptionskip}{0pt}
  \includegraphics[width=1.0\linewidth]{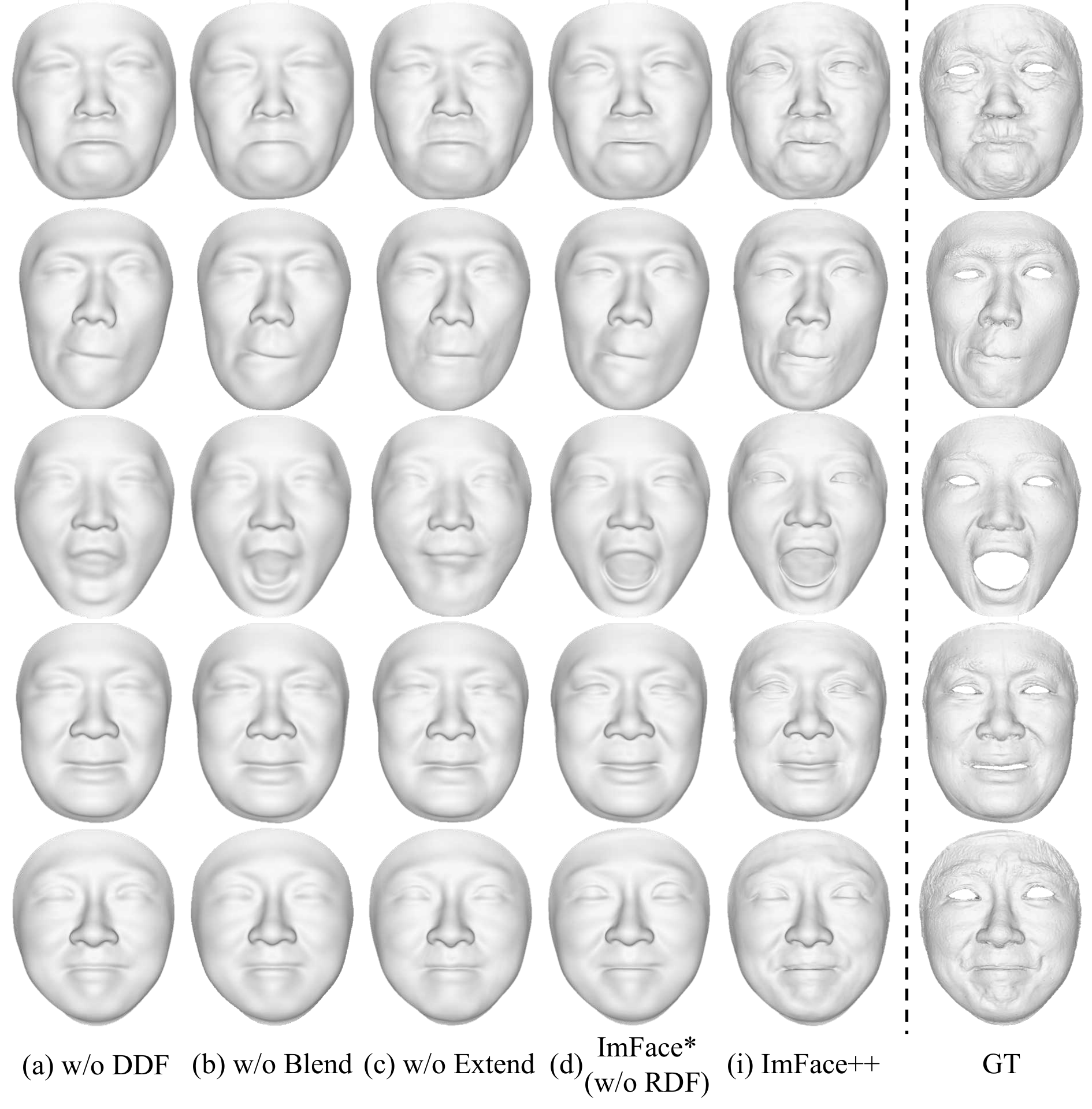}
   \caption{Qualitative ablation study results.}
   \label{fig:ablation}
\end{figure}

\textbf{On Disentangled Deformation Fields.} To underscore the significance of the disentangled deformation learning process, we establish a baseline network that incorporates only one deformation field to universally learn face shape morphs. In this configuration, ${\mathbf z}_{exp}$ and ${\mathbf z}_{id}$ are concatenated and used as the input for the hyper net. Fig.~\ref{fig:ablation} (a) visually demonstrates the results. Despite some fine-grained details introduced by other designs, there is noticeable chaos in the reconstructed faces, especially for those with exaggerated expressions. The quantitative results presented in Tab.~\ref{table:ablation} also indicate the significance of decoupled deformation learning.

\textbf{On Neural Blend-Field.} We substitute the Neural Blend-Field in $\mathcal{E}$, $\mathcal{I}$, and $\mathcal{T}$ with standard MLPs containing an equivalent number of parameters. These MLPs directly predict global deformations or SDF values for the entire face. As illustrated in Fig.~\ref{fig:ablation} (b), visible blurriness appears due to the limited capability to learn non-linear facial changes. The quantitative results in Tab.~\ref{table:ablation} further affirm the necessity of the Neural Blend-Field in achieving accurate representations.

\textbf{On Improved Embedding Learning.} This strategy is introduced to facilitate the learning of more diverse and fine-grained facial expression deformations. As demonstrated in Fig.~\ref{fig:ablation} (c), when the number of expression embeddings is constrained to be the same as the expression categories, the generated expressions tend to be average. Furthermore, for large expressions, such as mouth stretching, the counterpart models struggle to converge to a reasonable state.

\begin{figure}
  \centering
  \setlength{\abovecaptionskip}{0pt}
  \setlength{\belowcaptionskip}{0pt}
  \includegraphics[width=1.0\linewidth]{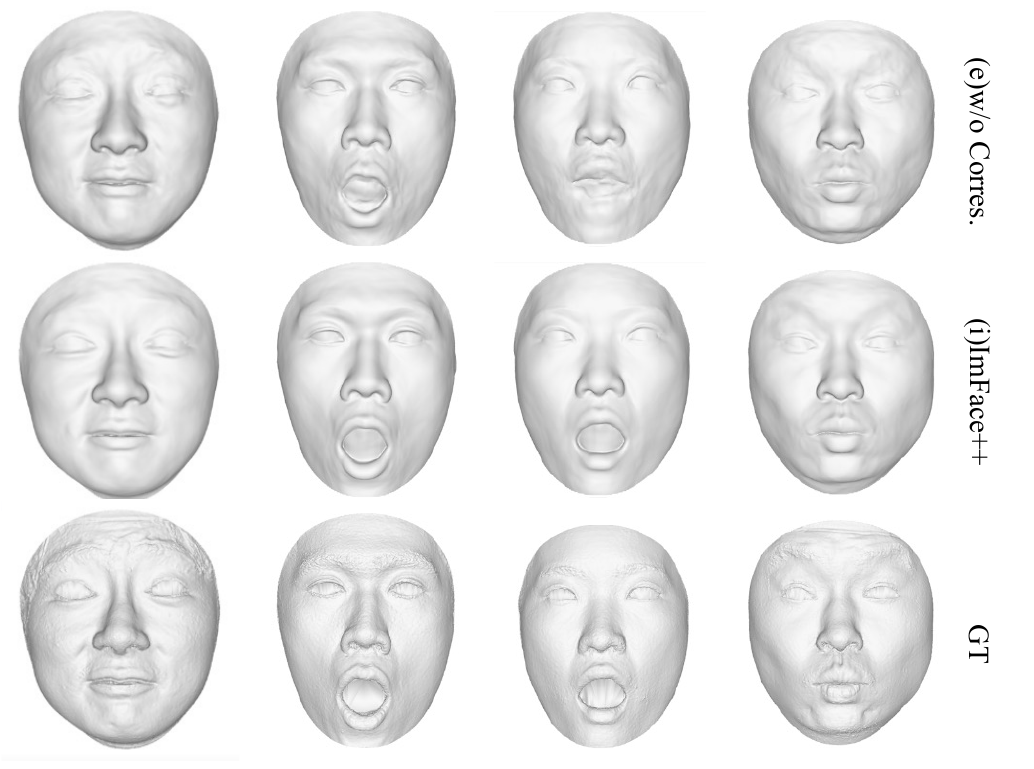}
   \caption{Illustration of ablation study results on point-to-point correspondences. Zoom in for a better view.}
   \label{fig:corres_ablate}
\end{figure}

\textbf{On Refinement Displacement Field.} RDF in ImFace++ is designed to learn and represent high-frequency details in facial shapes. Direct comparison between Fig.~\ref{fig:ablation} (d) and (i) illustrates that while ImFace can partially capture nonlinear facial deformations, it faces challenges in representing high-frequency details and personalized nuances. In contrast, ImFace++ excels at learning these intricate, subtle differences, by specially encoding them within the template space, exemplified by the sophisticated reproduction of features like the eyelid and wrinkles around the mouth. Furthermore, we conduct an experiment (w/o Corres.) to assess the impact of established correspondence on learning high-frequency details, where the displacement distance $\mathcal{D}:({\mathbf p_{b}}, {\mathbf z}_{detail}, l)  \mapsto  {\mathbf d} \in {\mathbb R}$ is modeled within the expression space instead of the aligned template space. The results are presented in Tab.~\ref{table:ablation} and Fig.~\ref{fig:corres_ablate}, which clearly indicate the presence of artifacts around the mouth and eyes.  This underscores the crucial role of the newly introduced RDF module and the impact of correspondence in ImFace++ for constructing a highly accurate 3D morphable model.

\textbf{On Hyper Net and Landmark Consistency Loss.} We conduct an ablation study with two variations of the HyperNetwork architecture:  
1) Concatenation (w/ Concat.): we concatenate the embeddings with coordinates and directly predict deformations and SDF using INRs.
2) Adaptive sentence normalization (w/ Ada.): We modify the vanilla implementation of adaptive sentence normalization by removing the normalization operation, enhancing the convergence with the sine activation function. 
As Tab.~\ref{table:ablation} shows, the Hyper Net architecture outperforms both variants, highlighting its efficacy in improving the model performance.
Furthermore, we ablate the landmark consistency loss (w/o ${\cal L}_{lmk_c}$). The results in Tab.~\ref{table:ablation} indicate that it leads to a better accuracy of facial features and its absence incurs noticeable degradation, underscoring its importance.

\begin{figure}
  \centering
  \setlength{\abovecaptionskip}{0pt}
  \setlength{\belowcaptionskip}{0pt}
  \includegraphics[width=1.0\linewidth]{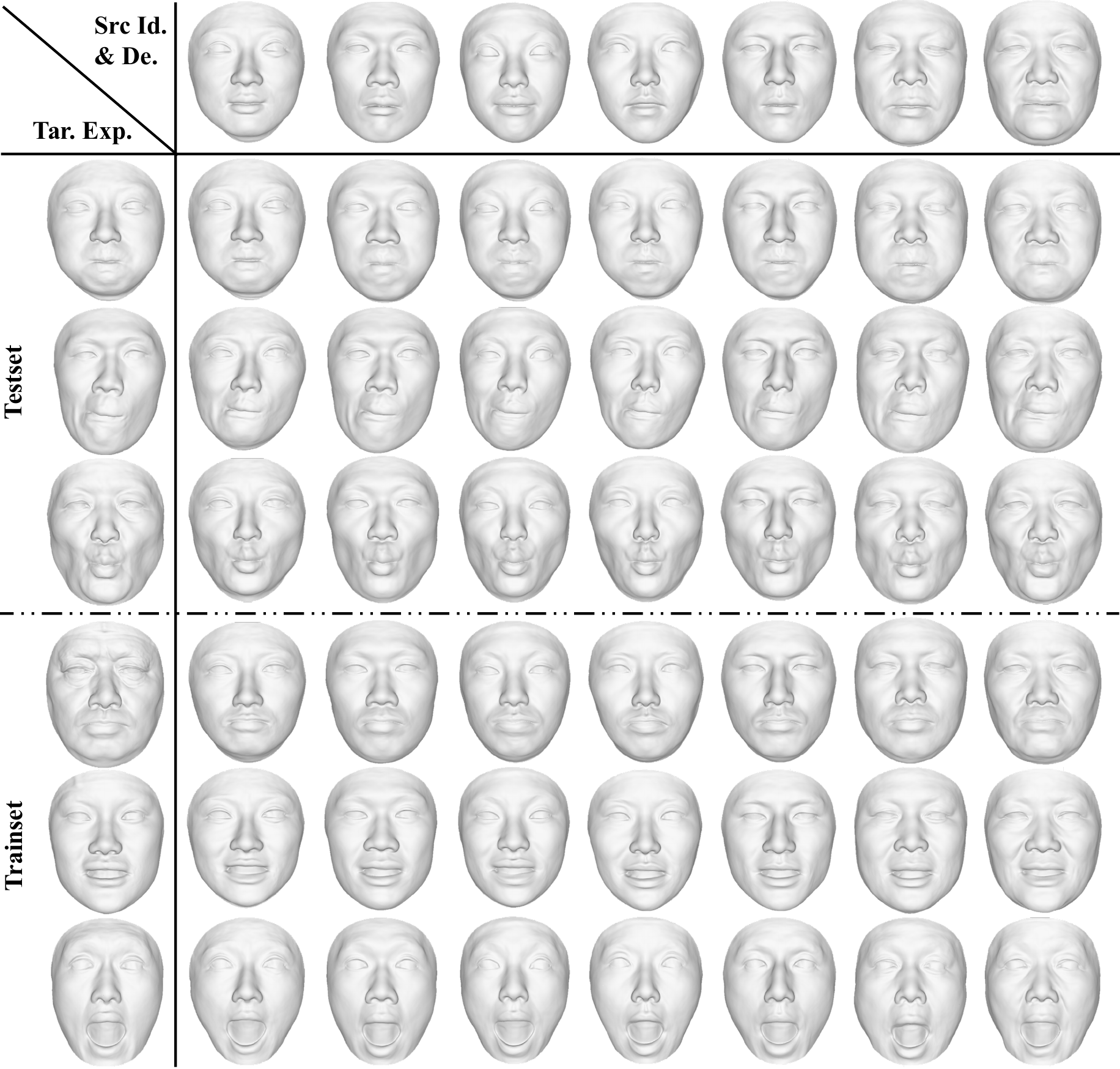}
   \caption{Expression editing results. The faces present in the first row provide the source identity and facial detail embeddings.}
   \label{fig:exp_edit}
\end{figure}

\subsection{Applications}

ImFace++ leverages embeddings as input, aligning with the input requirements of traditional 3DMMs. This enables ImFace++ to effectively perform a broad range of downstream tasks traditionally handled by conventional 3DMM methods, such as facial registration and other related applications~\cite{wu2021synergy, guo2020towards, li2023dsfnet}.

\subsubsection{Expression Editing}
Similar to existing 3DMMs, ImFace++ serves as a comprehensive facial representation built upon prior distributions of facial expression and identity morphs, making it suitable for a variety of downstream applications. In Fig.~\ref{fig:exp_edit}, we present expression editing results obtained from the FaceScape test set. The faces in the first row are the reconstructed ones providing source identities and neutral expressions, while the ones in the first column showcase the target expressions. In this figure, the first four rows depict reconstructed faces from the test set, and the last three rows feature the generated faces from the train set. It is evident that facial expression editing can be easily accomplished by modifying the expression embeddings. These lifelike 3D faces demonstrate the high representational capabilities of ImFace++.

\begin{figure}
  \centering
  \setlength{\abovecaptionskip}{0pt}
  \setlength{\belowcaptionskip}{0pt}
  \includegraphics[width=1.0\linewidth]{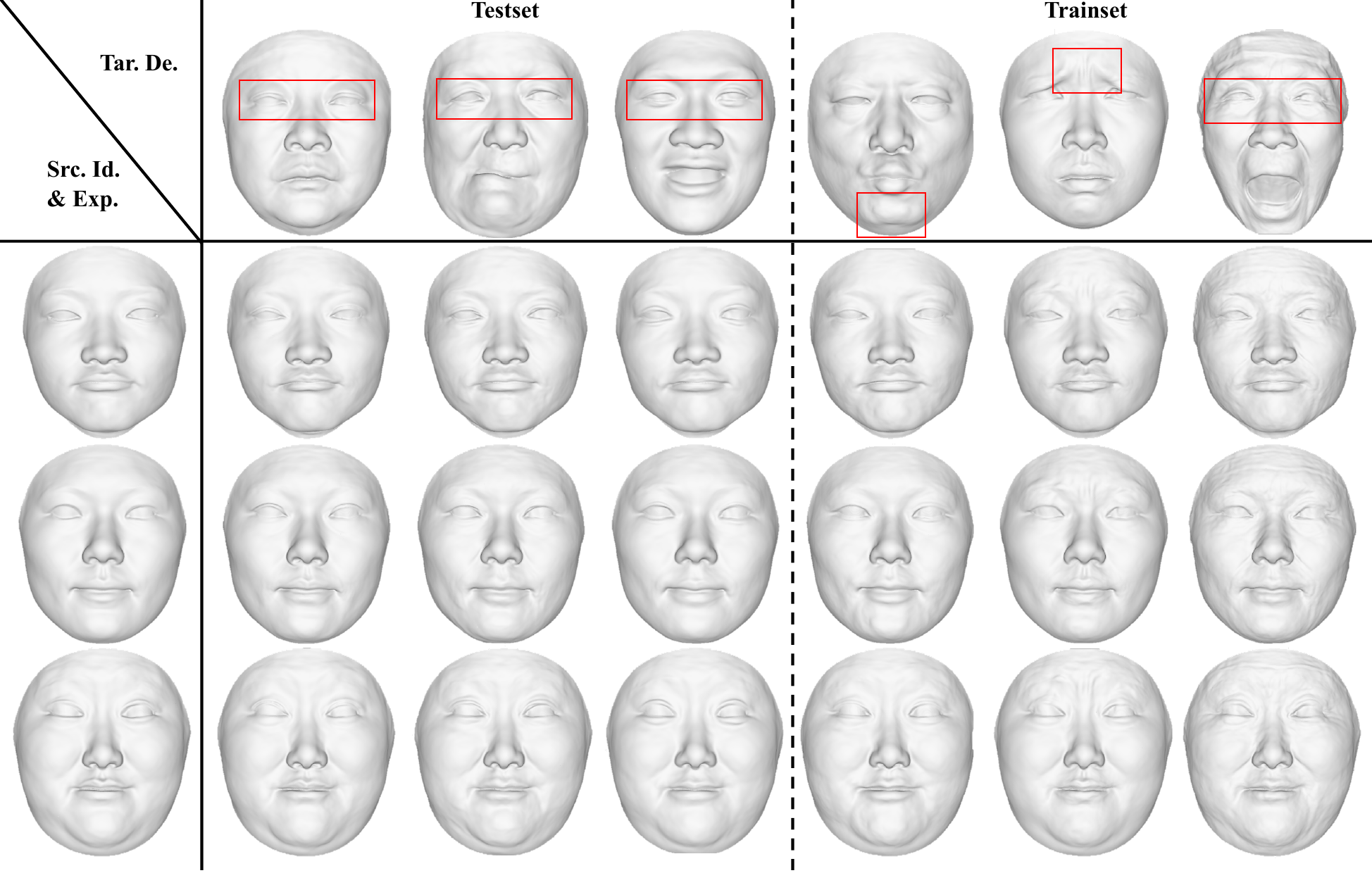}
   \caption{Facial detail transfer results. ImFace++ is able to edit high-frequency facial details by modifying the corresponding embeddings. The faces in the first column provide the source identity and expression embeddings. }
   \label{fig:detail_trans}
\end{figure}

\subsubsection{Facial Detail Transfer}
In Fig.~\ref{fig:detail_trans}, ImFace++ demonstrates its capability to transfer high-frequency facial details across different identities and expressions. In particular, the target faces in columns 2 to 4 exhibit various eyelid types, a challenging task for explicit face representations. ImFace++ accomplishes the transfer of high-frequency details with impressive fidelity by merely swapping the detail embeddings. Furthermore, the model can also transfer facial wrinkles, such as frown lines and crow's feet in the last two columns, to the source faces, showcasing its ability to capture and replicate fine-grained facial features with a remarkable precision.

\subsubsection{Texture Transfer}
\begin{figure}
  \centering
  \setlength{\abovecaptionskip}{0pt}
  \setlength{\belowcaptionskip}{0pt}
  \includegraphics[width=1.0\linewidth]{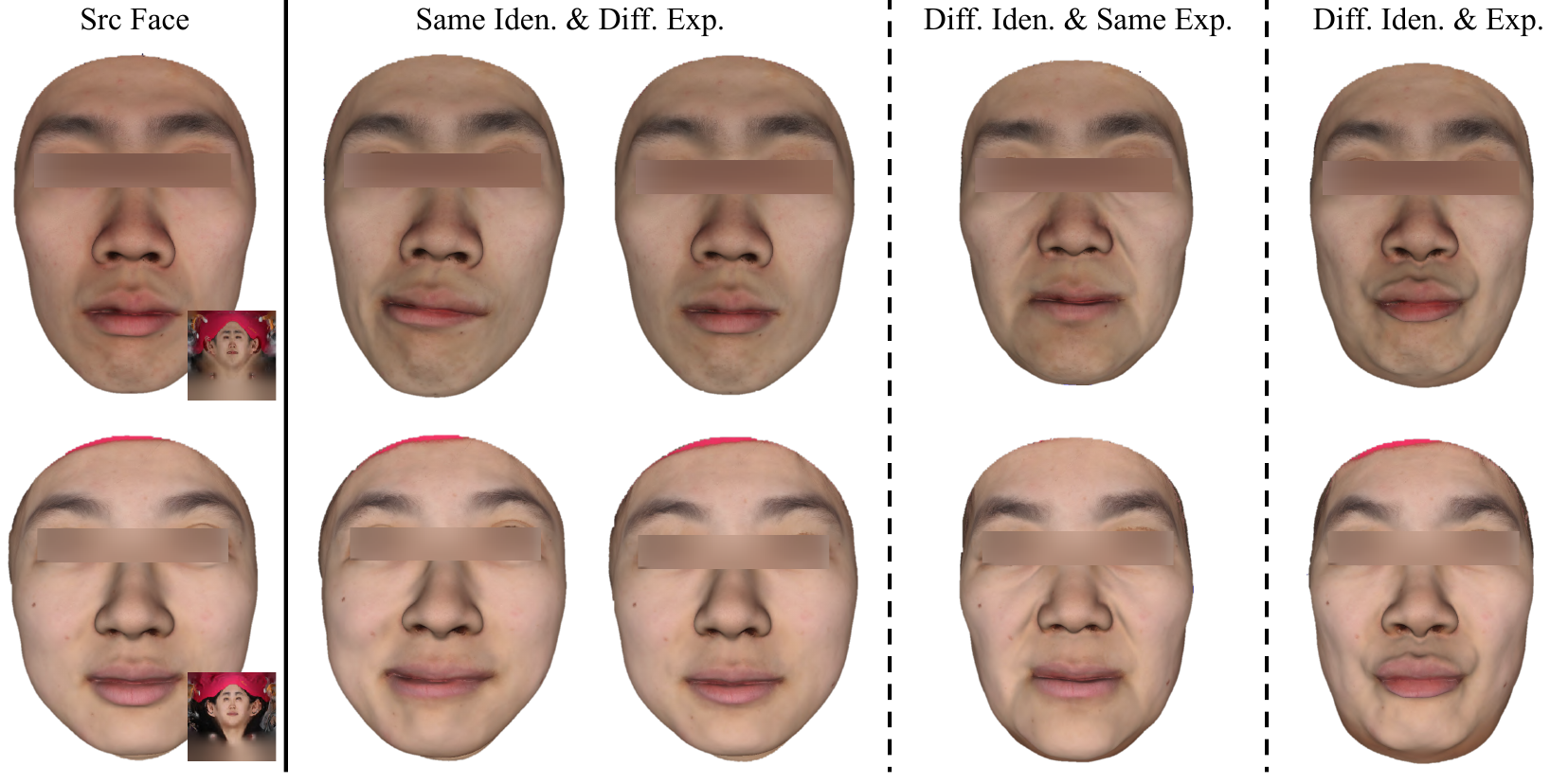}
   \caption{Illustration of texture transfer. Leveraging the established correspondence, ImFace++ effectively transfers facial textures among different scans. Eyes are blurred for privacy.}
   \label{fig:texture_corr}
\end{figure}

\noindent ImFace++ establishes high-quality correspondences, enabling effective texture transfer. As illustrated in Fig.~\ref{fig:texture_corr}, a UV texture map from the source face can be mapped to other target faces by utilizing the established correspondence between them, as described in Sec.\ref{correspond}. This application further validates the robustness and accuracy of the correspondence quality in ImFace++.

\subsubsection{Multi-view Reconstruction}

\begin{figure}
  \centering
  \setlength{\abovecaptionskip}{0pt}
  \setlength{\belowcaptionskip}{0pt}
  \includegraphics[width=1.0\linewidth]{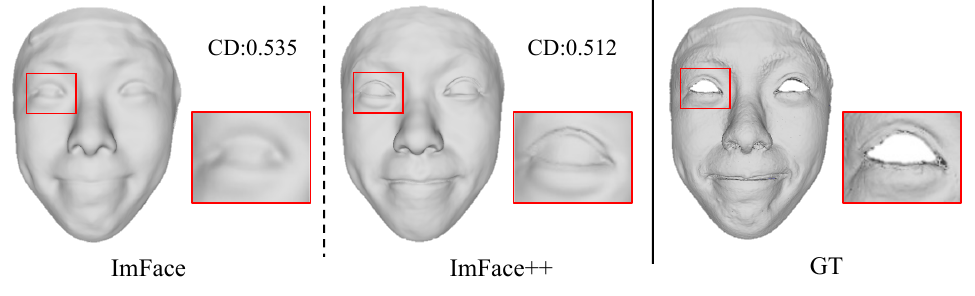}
   \caption{Multi-view reconstruction results. The Chamfer distance (CD) is listed at the right of each method.}
   \label{fig:mvs}
\end{figure}

\noindent  ImFace++ is fully differentiable, enabling us to employ neural rendering techniques for reconstructing facial shapes from multi-view images. In particular, we use NeuFace \cite{zheng2023neuface} as our rendering framework, utilizing either ImFace or ImFace++ as the geometry module. Fig.~\ref{fig:mvs} displays the recovered geometries. As can be seen, ImFace++ serves as a robust facial geometric prior model, enhancing ImFace by effectively capturing high-frequency features.

\begin{table}[htb]
  \centering
  \setlength{\belowcaptionskip}{0pt}
  \begin{tabular}{cc}
    \toprule
    Methods & Forwarding Time(s) \\
    \midrule
    i3DMM~\cite{yenamandra2021i3dmm} & 0.00552 \\
    NPHM~\cite{giebenhain2022learning} & 0.00603 \\
    ImFace~\cite{zheng2022imface} & 0.00456 \\
    ImFace++ & 0.01961 \\
    \bottomrule
  \end{tabular}
  \caption{Forwarding time comparison. We evaluate the time required to forward 8,192 points using different INR-based models on a single NVIDIA Tesla A100 GPU. Neural Blend Fields are re-implemented in parallel following \cite{nphm_github} to accelerate training and inference.}
  \label{table:inference_time}
\end{table}

\subsubsection{Ear-to-ear Face Reconstruction}

\begin{figure}
  \centering
  \setlength{\abovecaptionskip}{0pt}
  \setlength{\belowcaptionskip}{0pt}
  \includegraphics[width=1.0\linewidth]{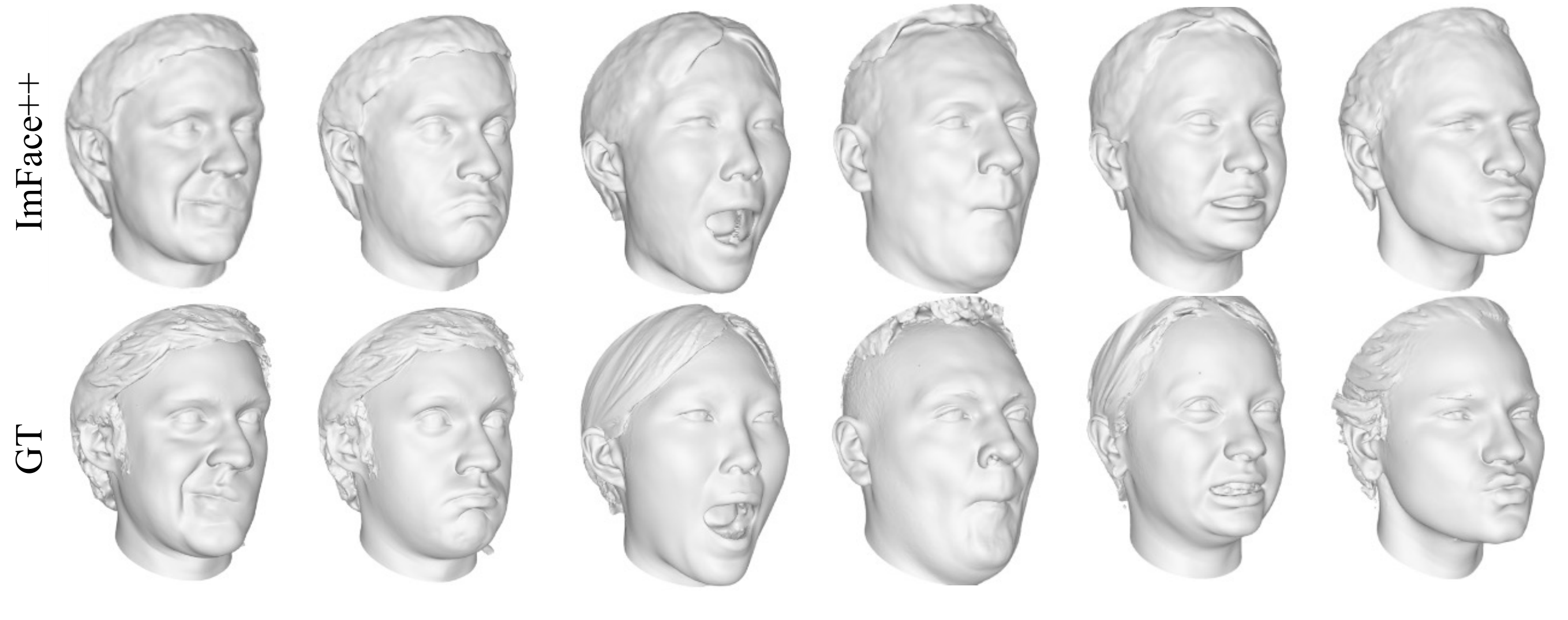}
   \caption{Reconstruction results on ear-to-ear facial shapes (trained on the NPHM dataset).}
   \label{fig:full_head}
\end{figure}

\noindent We utilize the NPHM~\cite{giebenhain2022learning} dataset to assess the performance of our method on ear-to-ear facial shapes. In the Neural Blend Field, we set $k=7$ to refine facial regions and enhance the capacity of ImFace++ for representing facial geometries. We follow the procedure in Sec. 4.1 to sample points for training. As illustrated in Fig.~\ref{fig:full_head}, ImFace++ can be seamlessly extended to render ear-to-ear facial shapes, achieving high-quality results across various expressions and identities. Additionally, ImFace++ effectively handles high-frequency details, such as wrinkles and eyelids.

\subsubsection{Texture Editing}

\begin{figure}
  \centering
  \setlength{\abovecaptionskip}{0pt}
  \setlength{\belowcaptionskip}{0pt}
  \includegraphics[width=1.0\linewidth]{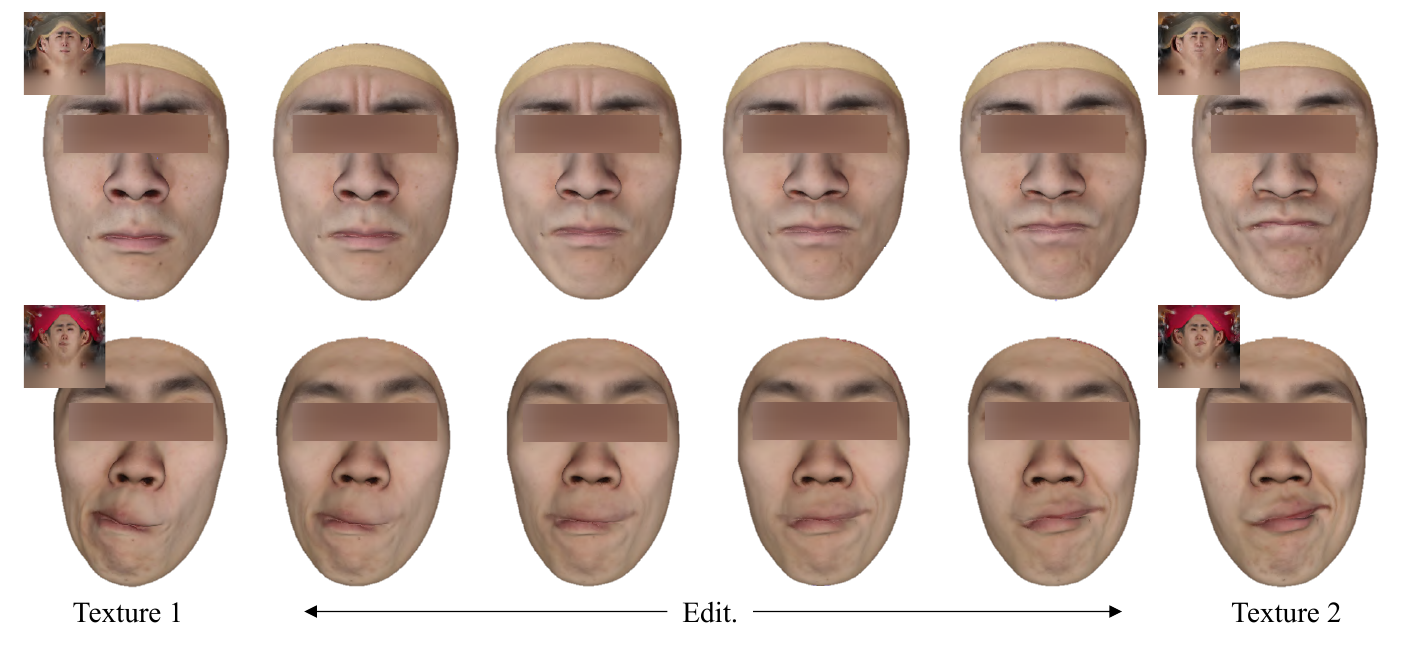}
   \caption{The illustration of texture editing results of ImFace++.}
   \label{fig:texture_editing}
\end{figure}

\noindent ImFace++ captures meaningful geometry clues and high-quality correspondences, enabling effective texture editing. Specifically, we perform interpolation in the embeddings and UV maps, utilizing the point-to-point correspondence to map the texture onto each face. As demonstrated in Fig.~\ref{fig:texture_editing}, the texture exhibits significant changes in accordance with modifications in facial geometry, such as the appearance of wrinkles around the brow and mouth.

\section{Discussion and Conclusion}

This paper introduces ImFace++, a sophisticated nonlinear 3D morphable face model with INRs. ImFace++ excels in capturing complex facial shape variations through the use of two explicitly disentangled deformation fields associated with expression- and identity-variations, along with a refinement displacement field that encodes high-frequency facial geometry details, contributing to the creation of highly realistic facial representations. To enhance the precision in capturing detailed facial deformations and geometries, ImFace++ also introduces a Neural Blend-Field. Furthermore, an improved embedding learning strategy is proposed that enables more fine-grained expressions. Additionally, the paper presents an effective preprocessing pipeline that enables INRs to work with non-watertight facial surfaces for the first time. Experimental results demonstrate that ImFace++ is highly competent and outperforms the state-of-the-art methods in this domain.

While this work represents a significant advancement in 3D facial shape representation, the focus is primarily on face geometry modeling, with less consideration given to facial appearances. Although a basic appearance model can be achieved by adding a color field or using UV color maps, there leaves space for further exploration in developing a more comprehensive INR-based 3DMM that accurately represents facial appearances by integrating higher quality editable UV maps. Alternatively, exploring more sophisticated appearance models would also offer promising avenues to enhance the realism and detail of facial textures. Addressing this issue is a significant direction for future research. Moreover, as shown in Tab.~\ref{table:inference_time}, INR-based methods generally require more computational resources and time for training and inference, as they necessitate forwarding neural networks for each point to calculate the SDF values. However, researchers can implement fully-fused MLPs under the tiny-cuda-nn framework \cite{tiny_github} or compute a cache of INRs \cite{garbin2021fastnerf} for faster training and inference. We leave this as our future work.

\vspace{-1mm}
\section*{Acknowledgments}
This work is partly supported by the National Key R\&D Program of China (2022ZD0161902), the National Natural Science Foundation of China (No. 62176012, 62202031, 62022011), Beijing Municipal Natural Science Foundation (No. 4222049), the Research Program of State Key Laboratory of Complex and Critical Software Environment, and the Fundamental Research Funds for the Central Universities.


%





%

\begin{IEEEbiography}[{\includegraphics[width=1in,height=1.25in,clip,keepaspectratio]{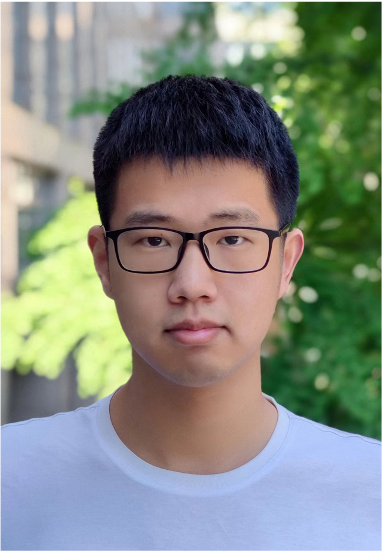}}]{Mingwu Zheng} received the B.E. and M.S. degree in computer science and technology from Beihang University, Beijing, China, in 2020 and 2023. His research interests include 3D neural rendering and generative modeling.
\end{IEEEbiography}

\begin{IEEEbiography}[{\includegraphics[width=1in,height=1.25in,clip,keepaspectratio]{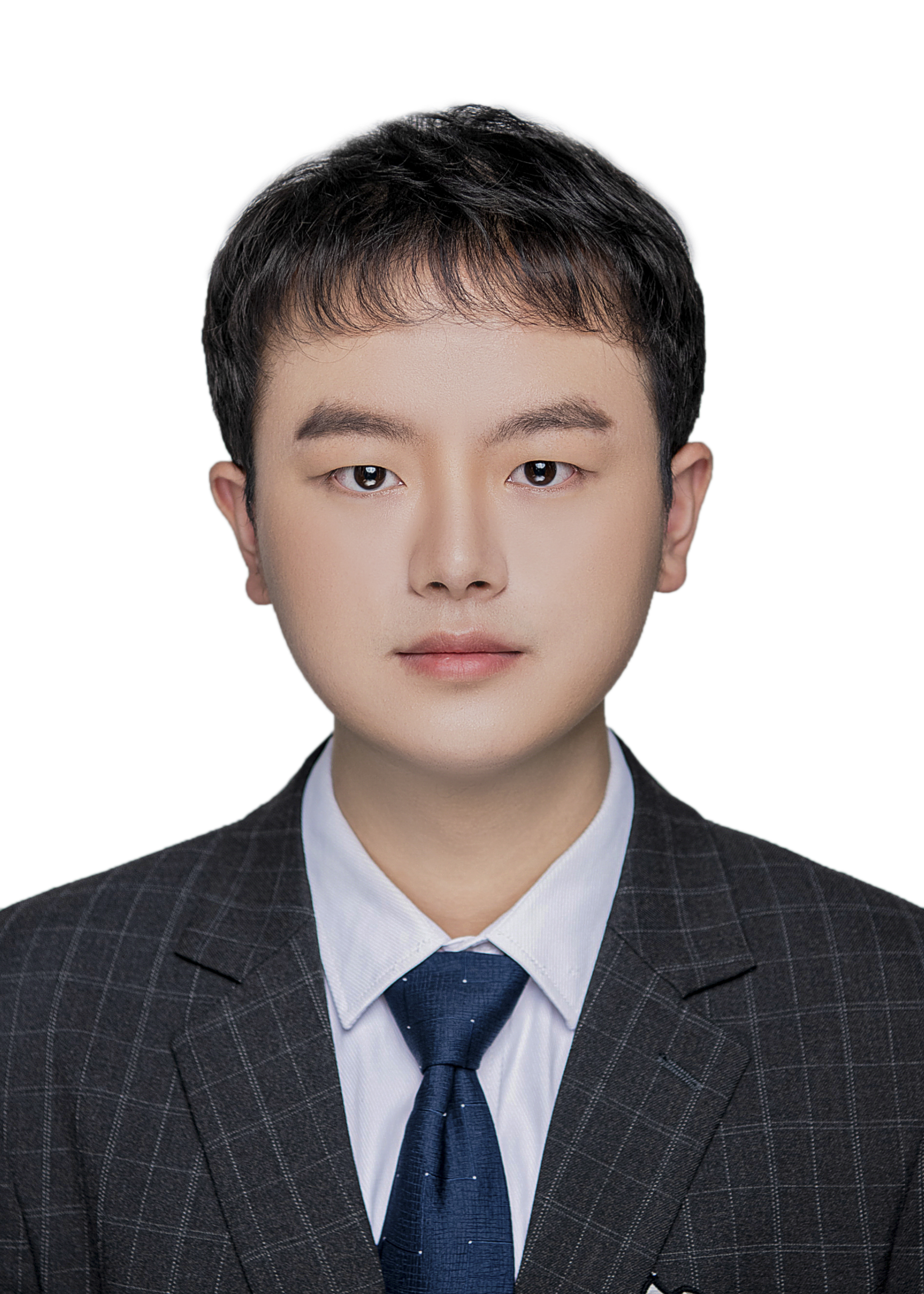}}]{Haiyu Zhang}
received the B.E. degree in computer science and technology from Beihang University, Beijing, China, in 2022, where he is currently working towards the Ph.D. degree with the Laboratory of Intelligent Recognition and Image Processing. His research interests include 3D face reconstruction, 3D face generation, and neural rendering.
\end{IEEEbiography}

\begin{IEEEbiography}[{\includegraphics[width=1in,height=1.25in,clip,keepaspectratio]{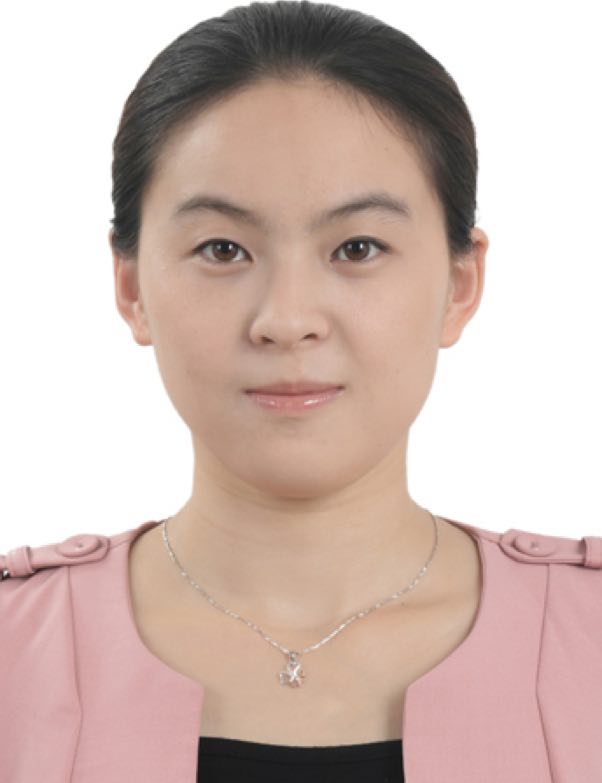}}]{Hongyu Yang}
received the B.E. and Ph.D. degrees in Computer Science and Technology
from Beihang University, Beijing, China, in 2013 and 2019, respectively. She is currently an associate professor at the School of Artificial Intelligence at Beihang University. Her research interests include image and video synthesis, as well as pattern recognition.
\end{IEEEbiography}

\begin{IEEEbiography}[{\includegraphics[width=1in,height=1.25in,clip,keepaspectratio]{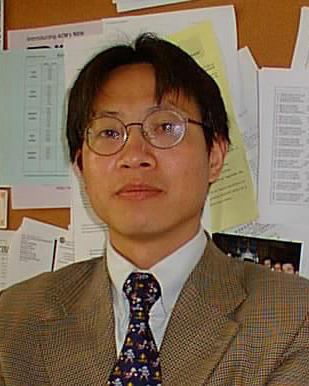}}]{Liming Chen} received the joint B.Sc. degree in mathematics and computer science from the University of Nantes, Nantes, France in 1984, and the M.Sc. and Ph.D. degrees in computer science from the University of Paris 6, Paris, France, in 1986 and 1989, respectively.

He first served as an Associate Professor with the Universit\'{e} de Technologie de Compi\`{e}gne, before joining \'Ecole Centrale de Lyon, \'Ecully, France, as a Professor in 1998,  where he leads an advanced research team on Computer Vision, Machine Learning and Multimedia. From 2001 to 2003, he also served as Chief Scientific Officer in a Paris-based company, Avivias, specializing in media asset management. In 2005, he served as Scientific Multimedia Expert for France Telecom R\&D China, Beijing, China. He was the Head of the Department of Mathematics and Computer Science, \'Ecole Centrale de Lyon from 2007 through 2016. His current research interests include computer vision, machine learning, image, and multimedia with a particular focus on robot vision and learning since 2016. Liming has over 300 publications and successfully supervised over 40 PhD students. He has been a grant holder for a number of research grants from EU FP program, French research funding bodies, and local government departments. Liming has so far guest-edited 5 journal special issues. He is an associate editor for Eurasip Journal on Image and Video Processing and a senior IEEE member.
\end{IEEEbiography}

\begin{IEEEbiography}[{\includegraphics[width=1in,height=1.25in,clip,keepaspectratio]{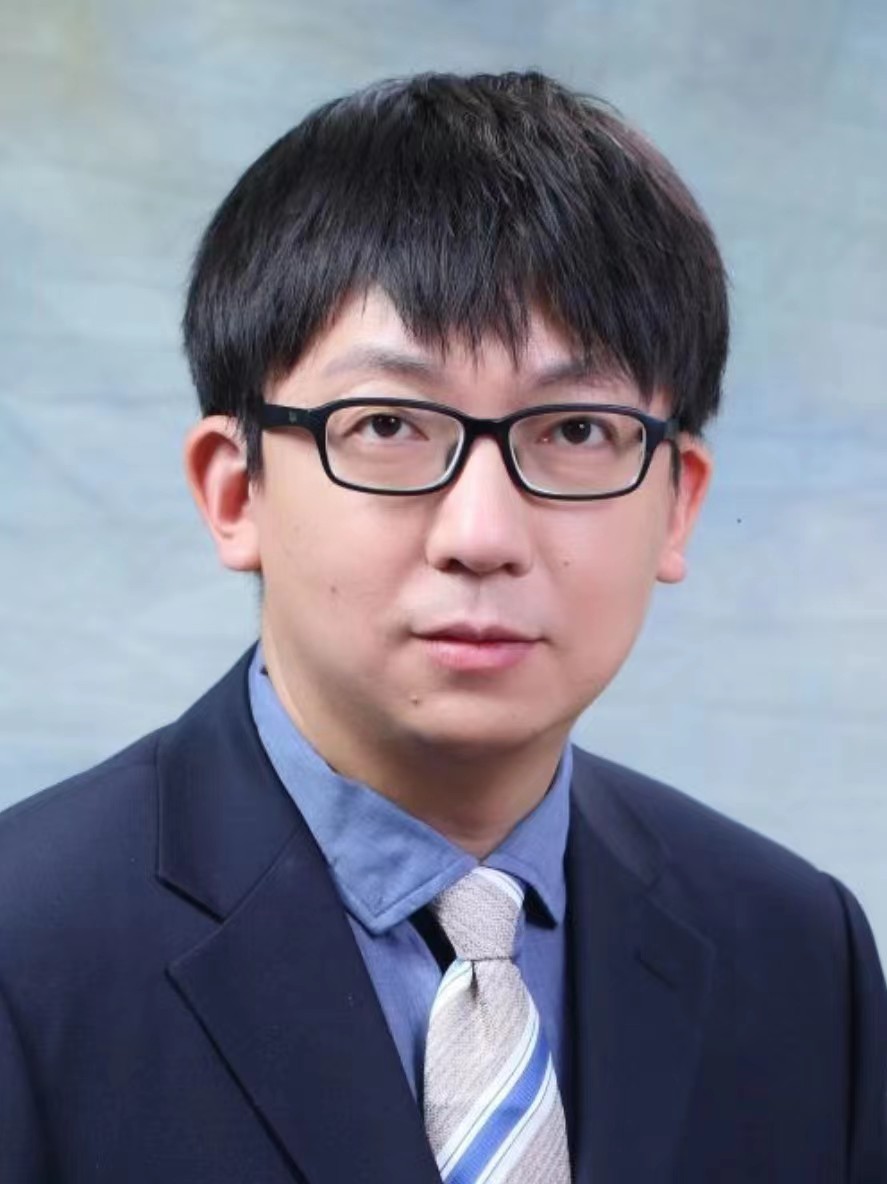}}]{Di Huang}
received the B.S. and M.S. degrees in computer science from
Beihang University, Beijing, China, in 2005 and 2008, respectively, and the Ph.D. degree in computer science from the \'Ecole Centrale de Lyon, Lyon, France, in 2011. He joined the Laboratory of Intelligent Recognition and Image Processing, School of Computer Science and Engineering, Beihang University, as a Faculty Member, where he is currently a Professor. His research interests include biometrics,
2D/3D face analysis, image/video processing, and pattern recognition.
\end{IEEEbiography}




\end{document}